\documentclass{article}

\usepackage{arxiv}
\usepackage{natbib}
\usepackage[utf8]{inputenc} % allow utf-8 input
\usepackage[T1]{fontenc}    % use 8-bit T1 fonts
\usepackage{hyperref}       % hyperlinks
\usepackage{url}            % simple URL typesetting
\usepackage{booktabs}       % professional-quality tables
\usepackage{amsfonts}       % blackboard math symbols
\usepackage{nicefrac}       % compact symbols for 1/2, etc.
\usepackage{microtype}      % microtypography
\usepackage{lipsum}
\usepackage{microtype}
\usepackage{graphicx}
\usepackage{subfigure}
\usepackage{booktabs} % for professional tables
\usepackage{amsfonts}       % blackboard math symbols
\usepackage{subfigure}
\usepackage{algorithm}
\usepackage{algorithmicx}
\usepackage{algpseudocode}
\usepackage{amsmath}
\usepackage{amsthm}
\usepackage{wrapfig}
\usepackage{setspace}
\usepackage{multirow}
\usepackage{tabularx}
\usepackage[table]{xcolor}
\newcommand{\norm}[1]{\left\lVert#1\right\rVert}
\DeclareMathOperator*{\argmax}{arg\,max}

\DeclareMathOperator*{\sign}{sign}
\setcitestyle{authoryear,round,citesep={;},aysep={,},yysep={;}}
\usepackage{hyperref}

\title{Guided Interpolation for Adversarial Training}

\author{
Chen Chen\textsuperscript{\rm 1}\thanks{Equal contributions.}$\;\,$, Jingfeng Zhang\textsuperscript{\rm 2}$^{*}$, Xilie Xu\textsuperscript{\rm 3}, Tianlei Hu\textsuperscript{\rm 1}, Gang Niu\textsuperscript{\rm 2} , Gang Chen\textsuperscript{\rm 1}\thanks{Corresponding author.}$\;\,$, Masashi Sugiyama\textsuperscript{\rm 2,4}\\
  \textsuperscript{\rm 1} College of Computer Science and Technology,
  Zhejiang University \\
  \textsuperscript{\rm 2}  RIKEN Center for Advanced Intelligence Project (AIP) \\
  \textsuperscript{\rm 3} Taishan College, Shandong University \\
  \textsuperscript{\rm 4}  The University of Tokyo \\
}

\begin{document}
\maketitle

\begin{abstract}
To enhance \textit{adversarial robustness}, \textit{adversarial training} learns deep neural networks on the \textit{adversarial variants} generated by their natural data. However, as the training progresses, the training data becomes less and less \textit{attackable}, undermining the robustness enhancement. 
A straightforward remedy is to incorporate more training data, but sometimes incurring an unaffordable cost.
In this paper, to mitigate this issue, we propose the \textit{guided interpolation framework} (GIF): in each epoch,  the GIF employs the previous epoch's meta information to guide the data's interpolation. Compared with the vanilla \textit{mixup}, the GIF can provide a higher ratio of attackable data, which is beneficial to the robustness enhancement; it meanwhile mitigates the model's linear behavior between classes, where the linear behavior is favorable to \textit{standard training for generalization} but not to \textit{adversarial training for robustness}. As a result, the GIF encourages the model to predict invariantly in the cluster of each class. 
Experiments demonstrate that the GIF can indeed enhance adversarial robustness on various adversarial training methods and various datasets. 

\end{abstract}

\section{Introduction}
Deep neural networks (DNNs) trained with a standard learning procedure are vulnerable to \textit{adversarial data}~\citep{biggio2013evasion,szegedy2013intriguing,goodfellow2014explaining,nguyen2015deep}. 
\textit{Adversarial training} (AT) is one of the most effective strategies for enhancing the model robustness against the adversarial data~\citep{Madry_adversarial_training}. To obtain the \textit{adversarial robustness}, AT methods alternatively generate adversarial data and optimize model parameters on the generated adversarial data~\citep{Madry_adversarial_training,Cai_CAT,Wang_Xingjun_MA_FOSC_DAT,wang2020improving_MART,wang2021on,zhang2019theoretically,zhang2020fat,Pang_ICML_19_AT_Ensemble,Pang2020Rethinking,pang2021bag,wu2020adversarial,babu_2020_CVPR,wong2020fast_zico_kolter,rice2020overfitting,babu_neurips_2020,bai2021improving_CAS}.

Recent studies on AT suggest the unequal treatment of data~\citep{ding2020mma,Wang_Xingjun_MA_FOSC_DAT,wang2020improving_MART,zhang2021geometry}. 
In particular, \citet{zhang2021geometry} divided the training data into two categories---\textit{attackable data} and \textit{guarded data}, in which attackable/guarded data are close to/far away from the class boundary that can/cannot be attacked. To enhance adversarial robustness, attackable data are particularly useful in learning the decision boundary ~\citep{wang2020improving_MART,zhang2021geometry}.

\begin{figure*}[!tp]
	\centering
	\subfigure{
		\includegraphics[width=0.38\linewidth]{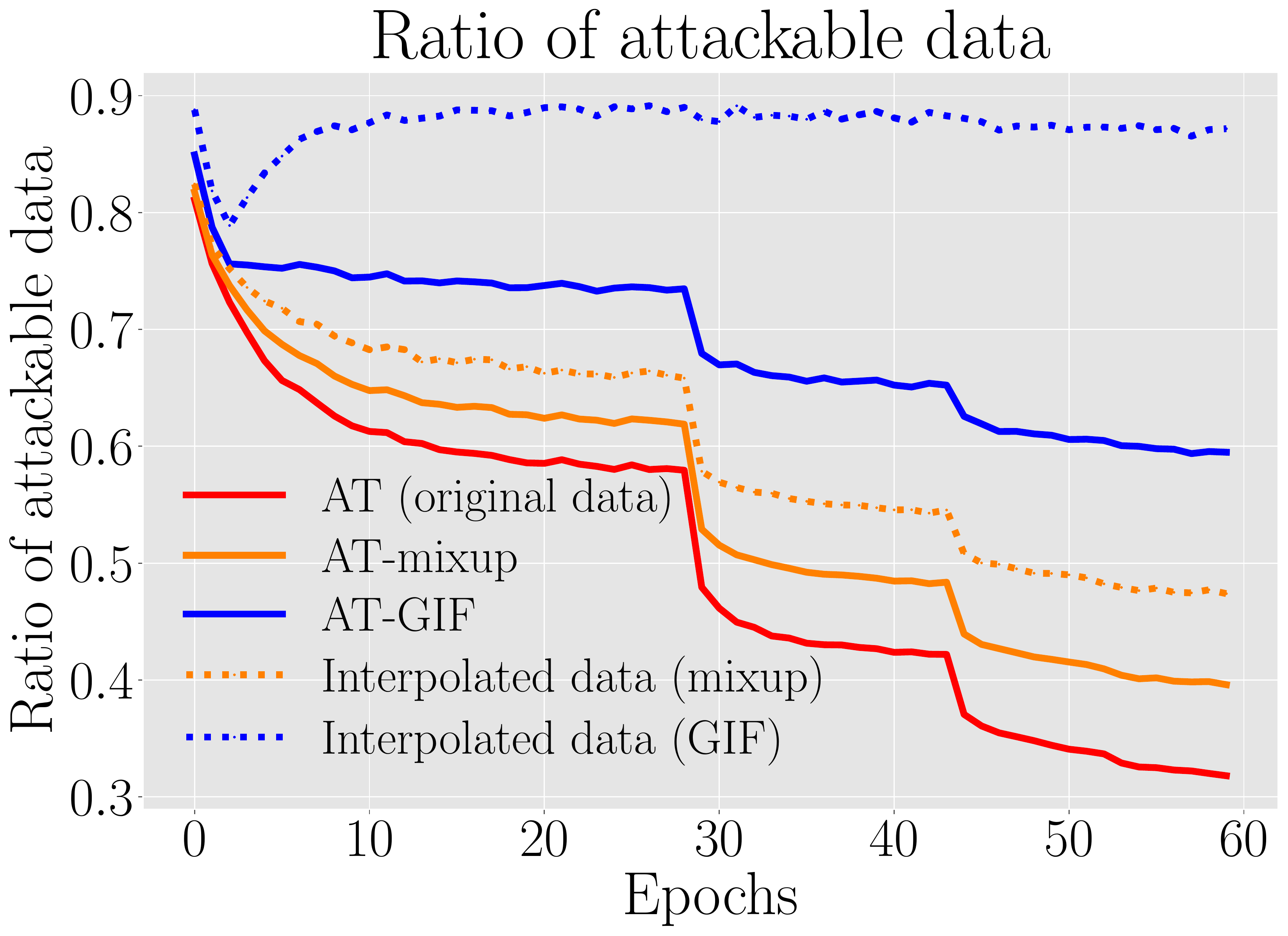}
	}
	\hspace{5mm}
	\subfigure{
		\includegraphics[width=0.38\linewidth]{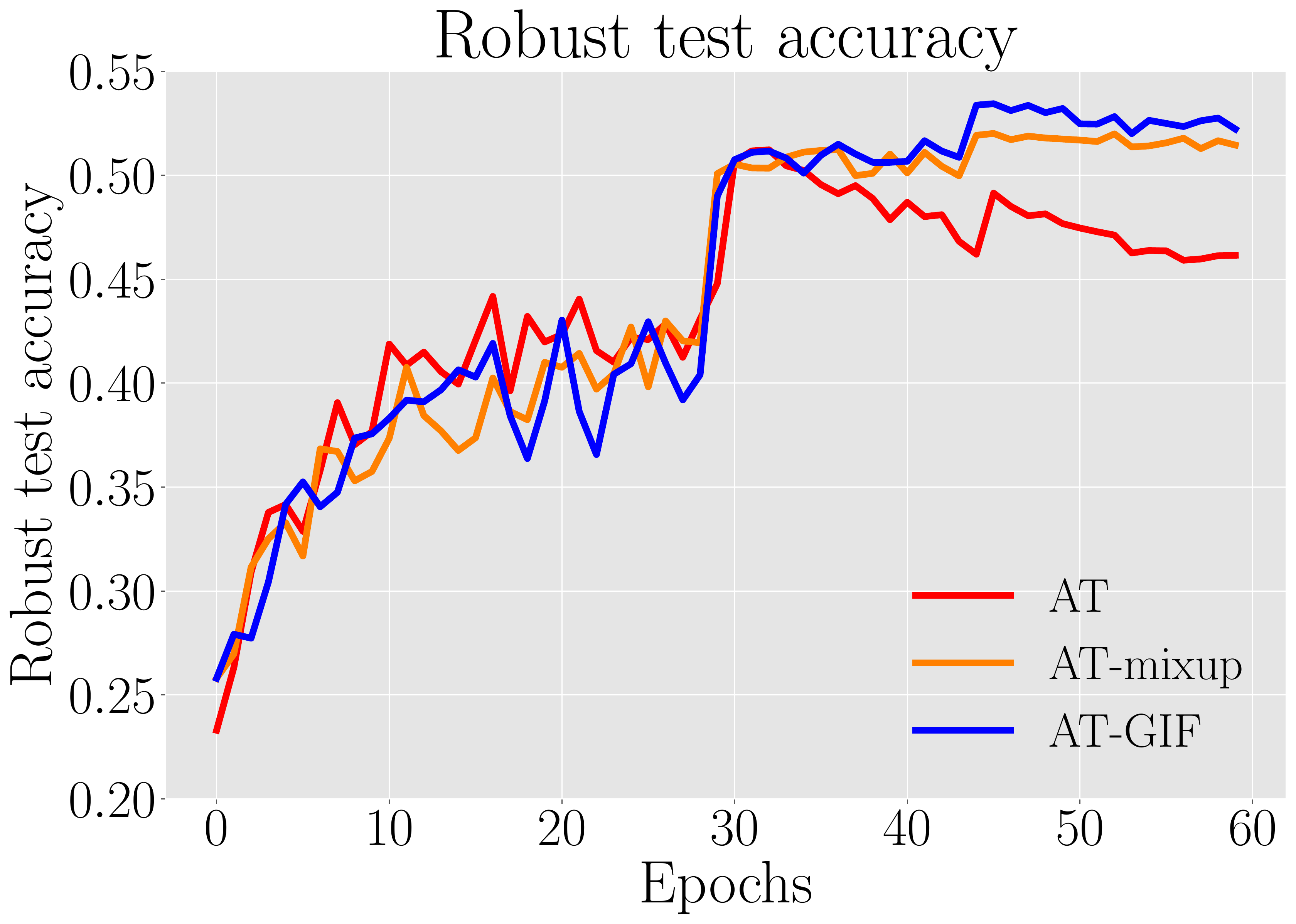}
	}
	%\vspace{-2mm}
	\caption{Comparisons between AT~\citep{Madry_adversarial_training}, AT-mixup (AT with vanilla mixup), and AT-GIF (AT with our GIF) on the CIFAR-10 dataset~\citep{krizhevsky2009learning}. %AT-mixup and AT-GIF train with adversarial data generated by original data and interpolated data. 
	The detailed experimental settings are in Appendix \ref{app_sec:fig12}. 
	\textbf{Left panel}: The attackable data ratio of AT (red line) decreases rapidly as the training progresses. 
	In each mini-batch, both AT-mixup and AT-GIF use $50\%$ original data and $50\%$ interpolated data for adversarial training. 
	Among the interpolated data (dotted lines), nearly 90\% by the GIF (blue dotted line) are attackable, while the interpolated data by vanilla mixup (orange dotted line) contain a much lower ratio of attackable data.
	Therefore, compared with AT-mixup (orange solid line), AT-GIF (blue solid line) can leverage a higher ratio of attackable data per epoch.
	\textbf{Right panel}: The robust test accuracy is evaluated under PGD-20 attack. The robust test accuracy of AT (red line) rises when the attackable data ratio is high (before Epoch 30). After Epoch 30 (with a reduced learning rate), the robust test accuracy drops significantly, which is strongly related to the low ratio of attackable data.
	Compared with AT-mixup (orange solid line), AT-GIF (blue solid line) has a higher robust accuracy.
	}
	\label{fig:att_acc}
	%\vspace{-2mm}
\end{figure*}

However, as the AT progresses, the ratio of attackable data decreases significantly, which jeopardizes the enhancement of adversarial robustness. 
In Figure~\ref{fig:att_acc}, we plot the ratio of attackable data in the left panel and the robust accuracy in the right panel.
The red lines show the training dynamics of a typical AT method~\citep{Madry_adversarial_training}. 
As the training progresses, more and more training data become guarded; thus, the ratio of attackable data decreases. After Epoch 30 (with a reduced learning rate), this ratio drops rapidly, while the robust accuracy ceases to rise but begins to drop. 
This strong correlation between this ratio and the robustness urges us to introduce more attackable data for AT.

A straightforward remedy for the shortage of attackable data is to incorporate more training data~\citep{schmidt2018adversarially}. 
\citet{schmidt2018adversarially} and \citet{hendrycks2019using} showed that AT for learning a robust model requires substantially more training data than \textit{standard training} (ST). Nevertheless, gathering additional data especially with high-quality labels is often expensive; therefore, \citet{carmon2019unlabeled}, \citet{DeepMind_useto}, and \citet{najafi2019robustness} leveraged a massive amount of unlabeled data for training adversarially robust models. However, in many privacy-sensitive applications such as medicine~\citep{buch2018artificial_medicine} and finance~\citep{abbe2012privacy_finance}, collecting massive unlabeled data is very costly and sometimes not possible at all~\citep{10.5555/3152676_GPPR}. 

To mitigate this shortage, this paper proposes a novel data augmentation framework---the \textit{guided interpolation framework} (GIF)---for AT. Without collecting massively additional data, the GIF employs the meta information of the previous epoch to guide the data's interpolation, and then, the interpolated data are used for the current-epoch training (details in Section \ref{sec:GIF_detail}). 
As shown in Figure~\ref{fig:att_acc}, compared with vanilla-mixup AT (AT-mixup, orange line), the AT with GIF (AT-GIF, blue line) has a higher ratio of attackable data, leading to a higher robust accuracy. 

In particular, the GIF is carefully designed for AT, which mitigates the \textit{model's linear behavior between classes}.
The model's linear behavior is introduced by the vanilla mixup~\citep{zhang2018mixup}; it encourages the model predictions that transit linearly from class to class, providing a smooth change of prediction confidence~\citep{zhang2018mixup} (two middle panels of Figure~\ref{fig:linear_behavior}). 
However, this linear behavior is favorable to ST for the generalization (and weak robustness against FGSM attack~\citep{zhang2021how}) but not to AT for the strong robustness against the adaptive attacks. 
AT encourages the model to be locally constant within the input's neighborhood~\citep{papernot2016towards,goodfellow2014explaining}, thus encouraging invariant predictions within the cluster of each class (two left panels of Figure~\ref{fig:linear_behavior}). 
To this end, the GIF mitigates defect of vanilla mixup (i.e., introducing the linear behavior) and encourages invariant predictions (two right panels of Figure~\ref{fig:linear_behavior}); it meanwhile can generate more attackable data, thus further enhancing the robustness. 

We summarize our contributions as follows.
\vspace{-3mm}
\begin{itemize}
    \item Although the vanilla mixup~\citep{zhang2018mixup} can indeed generate more attackable data (compared with AT), we show that it meanwhile introduces the model's linear behavior between classes, which is never our desideratum for adversarial robustness (Section \ref{sec:Motivation}).
    \item We design a novel guided interpolation framework (GIF) for adversarial training (AT). To enhance adversarial robustness, the GIF can mitigate the undesirable linear behaviors and meanwhile provide more attackable data (Section~\ref{sec:GIF}). Moreover, the GIF is a compatible approach: it can be easily incorporated into existing adversarial training methods, e.g., TRADES~\citep{zhang2019theoretically} and GAIRAT~\citep{zhang2021geometry}.
    \item Experiments on various datasets and adversarial training methods corroborate the efficacy of our GIF in enhancing adversarial robustness (Section \ref{sec:Exp}).
    %We identify the side effect of the vanilla mixup for the adversarial robustness, i.e., introducing the model's linear behavior between classes .
\end{itemize}

\section{Preliminary and Related Work}
%In this section, we review adversarial training and data interpolation techniques.
Section~\ref{section:Review_AT} reviews a typical AT method, i.e., standard adversarial training by \citet{Madry_adversarial_training}, and reviews recent other AT methods that claimed the unequal treatment of data. 
%show that standard AT gets overfitted due to the low ratio of attackable data.
Section~\ref{sec:Review_mixup} reviews the mixup~\citep{zhang2018mixup} and its relation with robustness. 
We also highlight the difference between our GIF and existing studies. 

\subsection{Adversarial Training}
\label{section:Review_AT}
Let $(x,d_{\infty})$ be the input feature space $\mathcal{X}$ with the inifinity distance metric $d_\infty(x,x')=\norm{x-x'}_\infty$, and $\mathcal{B}_\epsilon(x)=\left\{x'\in\mathcal{X}|d_\infty(x,x')<\epsilon\right\}$ be the closed ball of radius $\epsilon>0$ centered at $x$ in $\mathcal{X}$. Let $S=\left\{(x_i,y_i)\right\}_{i=1}^n$ be a dataset, where $x_i\in \mathcal{X}$ is the input, and $y_i\in\mathcal{Y}$ is the one-hot label of $x_i$. Let $C$ be the number of classes, $f^c(\cdot)$ be the $c$-th element of the model's prediction $f(\cdot)$, and $y^c$ be the $c$-th element of one-hot label $y$.
\subsubsection{Standard Adversarial training}
\label{sec:at}
The objective function of standard adversarial training (AT)~\citep{Madry_adversarial_training} is 
\begin{equation}\label{eq:at}
\begin{split}
\min_{f_\theta\in\mathcal{F}}\frac{1}{n}\sum_{i=1}^n\ell(f_\theta(\tilde{x}_i),y_i),
\end{split}
\end{equation}
where
\begin{equation}\label{eq:at_adv_x}
\begin{split}
\tilde{x}_i=\argmax_{\tilde{x}\in\mathcal{B}_\epsilon(x_i)}\ell(f_\theta(\tilde{x}),y_i).
\end{split}
\end{equation}
$\tilde{x}_i$ is the most adversarial data within the $\epsilon$-ball centered at $x$. $f_\theta(\cdot):\mathcal{X}\rightarrow\mathbb{R}^C$ is a score function with parameters $\theta$. 
$\ell:\mathbb{R}^C\times\mathcal{Y}\rightarrow\mathbb{R}$ is the loss function (e.g., the cross-entropy loss).
Adversarial training can be divided into two steps: The first step maximizes the loss to find adversarial data; the second step minimizes the loss on the adversarial data w.r.t. the parameters $\theta$.

Standard AT uses projected gradient descent (PGD) to approximate Eq. (\ref{eq:at_adv_x}) and generate adversarial data. Given a starting point $x^{(0)}\in\mathcal{X}$, step size $\alpha>0$, radius $\epsilon>0$, and maximum number of steps $K$, PGD searches adversarial data as follows:
\begin{equation}\label{eq:pgd}
\begin{split}
x^{(k+1)}=\Pi_{\mathcal{B}_\epsilon(x^{(0)})}\left(x^{(k)}+\alpha\sign(\nabla_{x^{(k)}}\ell(f_\theta(x^{(k)}),y))\right),
\end{split}
\end{equation}
where $k=0, \ldots, K-1$ is the step number, $x^{(0)}$ refers to natural data or natural data perturbed by a small Gaussian or uniform random noise, $y$ is the corresponding label for the natural data, $x^{(k)}$ is the adversarial data at step $k$, and $\Pi_{\mathcal{B}_\epsilon(x^{(0)})}(\cdot)$ projects the adversarial data onto the $\epsilon$-ball centered at $x^{0}$. Eq. (\ref{eq:pgd}) increases the loss by gradient ascent and then projects the updated adversarial data back onto the $\epsilon$-ball centered at $x^{0}$.

After $K$ steps of PGD, standard AT utilizes the generated adversarial data $x^{K}$ to update the model parameters $\theta$.
On evaluation, the \textit{robust accuracy} is used, which is the fraction of predictions that the model correctly makes on the adversarial data generated by the adversarial attacks such as the PGD attack (in Eq.~\eqref{eq:pgd}).
%We denote the accuracy when attacked by a PGD adversary as the \textit{robust accuracy}.

\begin{figure*}[ht!]
  \centering
  \subfigure{
		\includegraphics[width=0.3\linewidth]{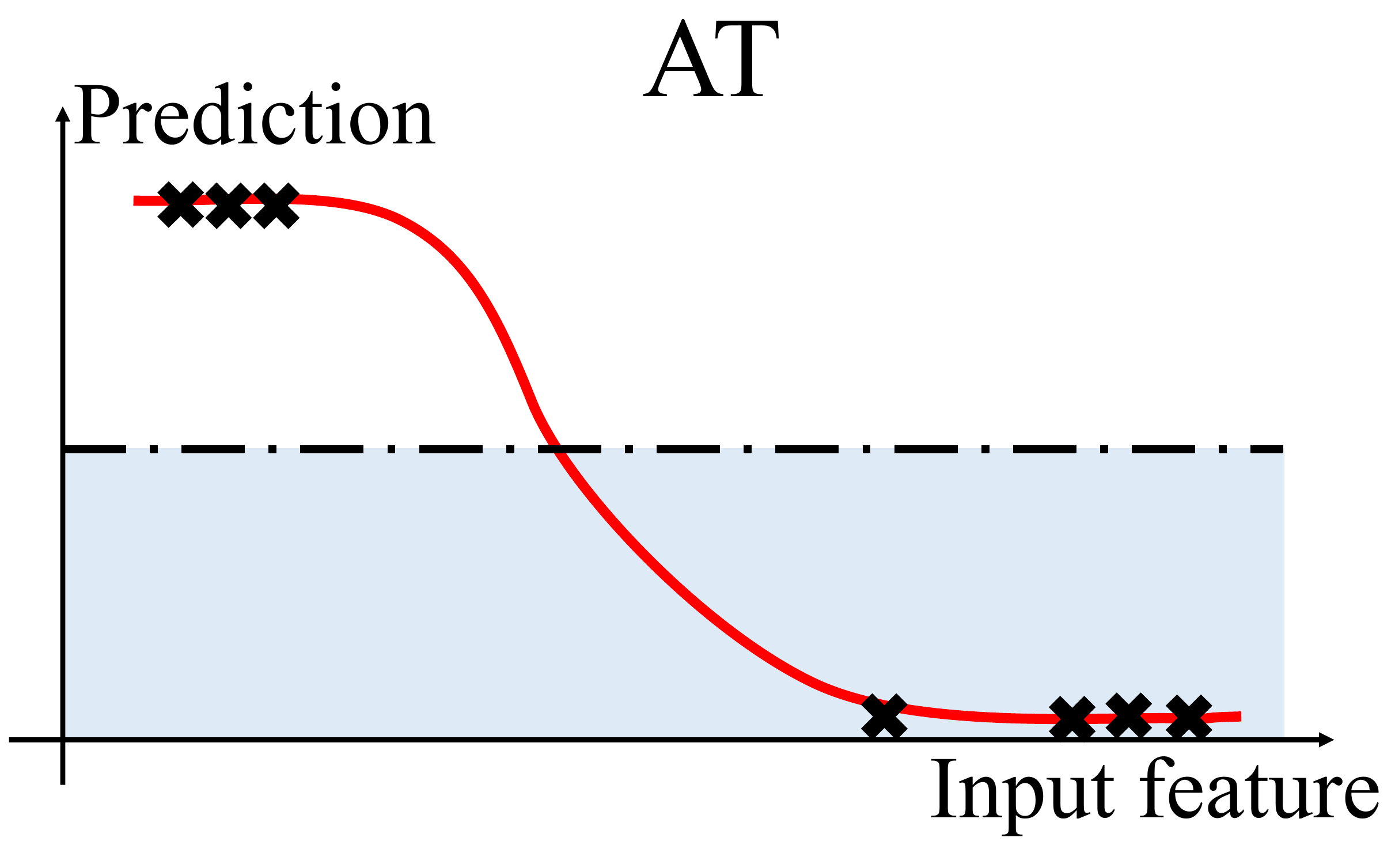}
	}
	\subfigure{
		\includegraphics[width=0.3\linewidth]{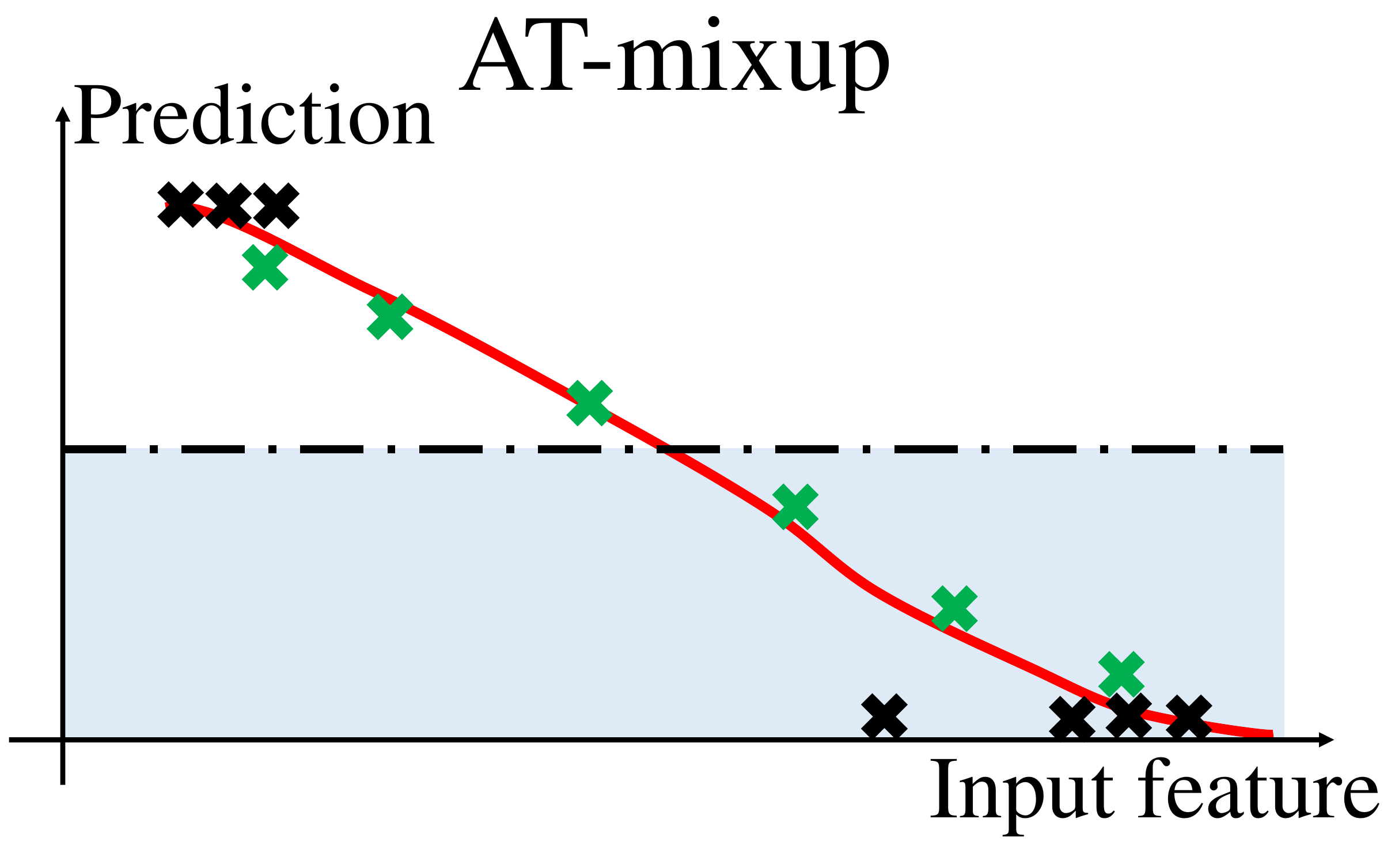}
	}
% 	\hspace{9mm}
    \subfigure{
		\includegraphics[width=0.3\linewidth]{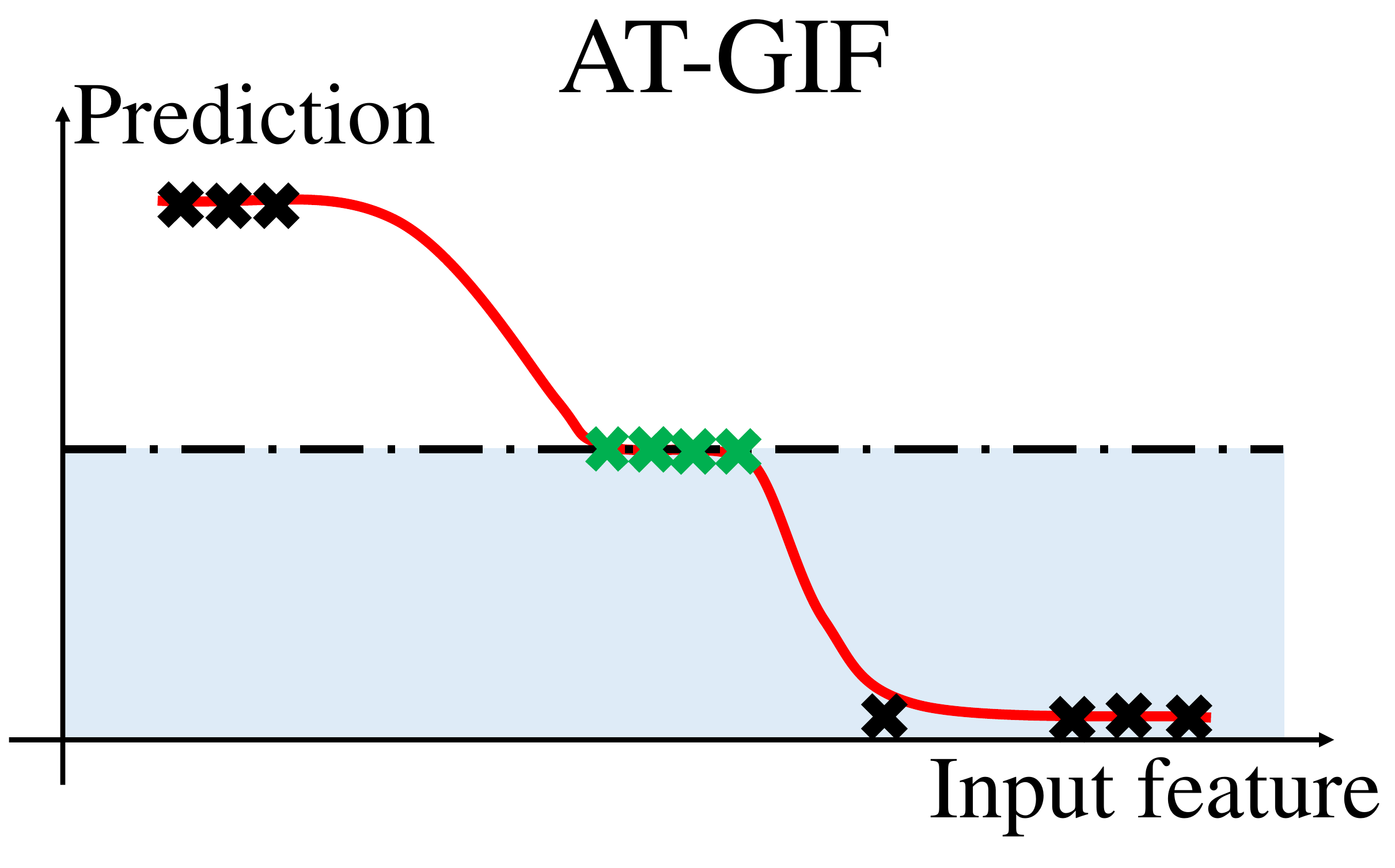}
	}
	\\
    \subfigure{
		\includegraphics[width=0.3\linewidth]{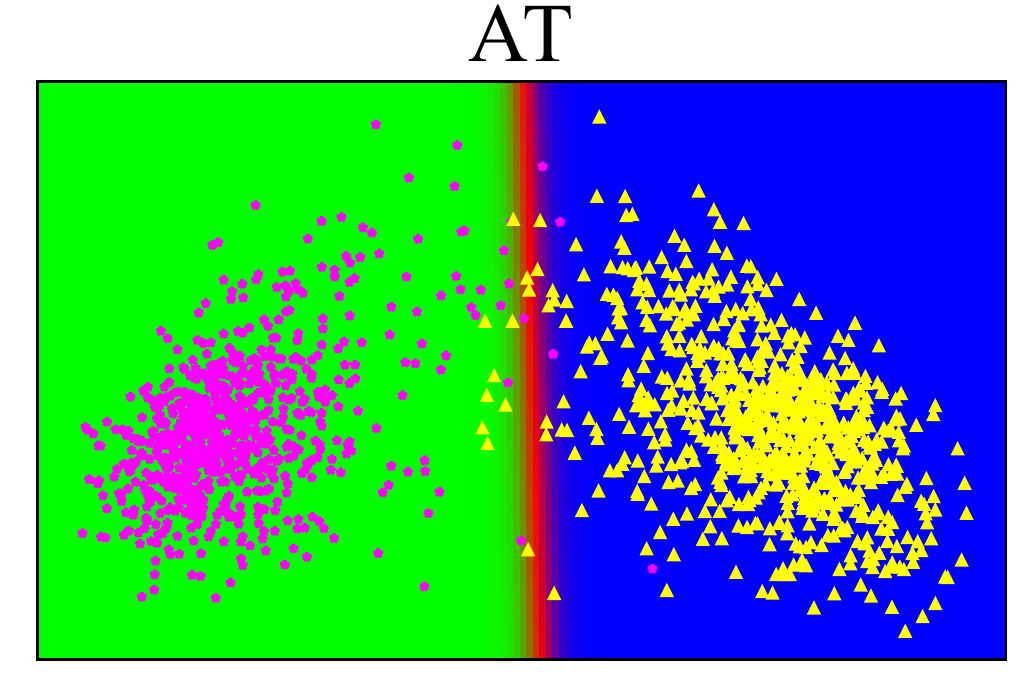}
	}
% 	\hspace{9mm}
    \subfigure{
		\includegraphics[width=0.3\linewidth]{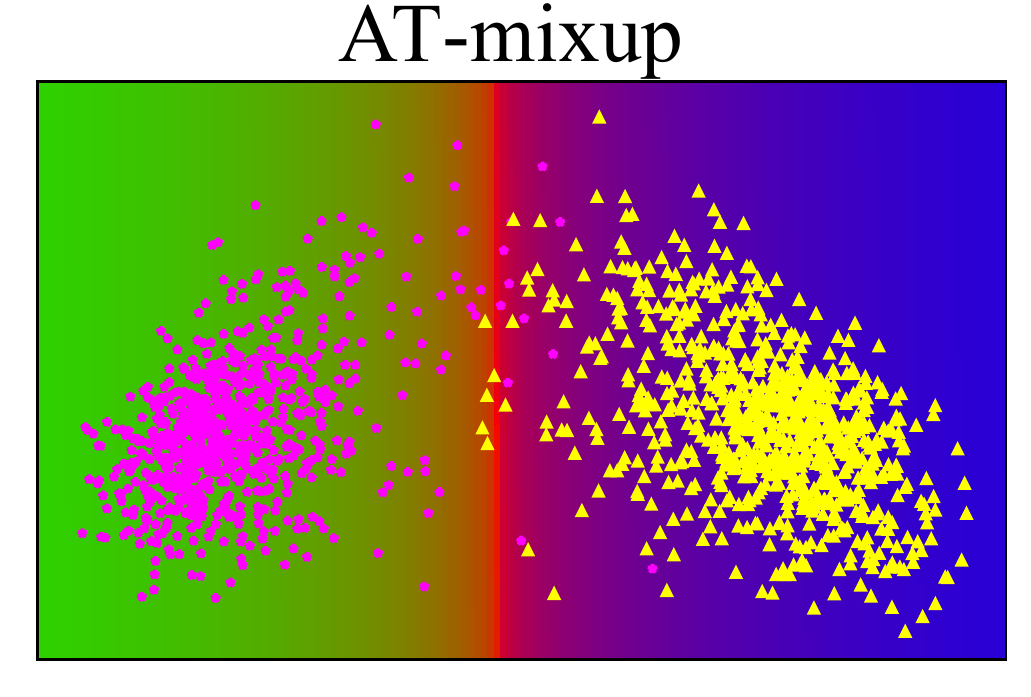}
	}
	%\hspace{2mm}
% 	\hspace{9mm}
	\subfigure{
		\includegraphics[width=0.3\linewidth]{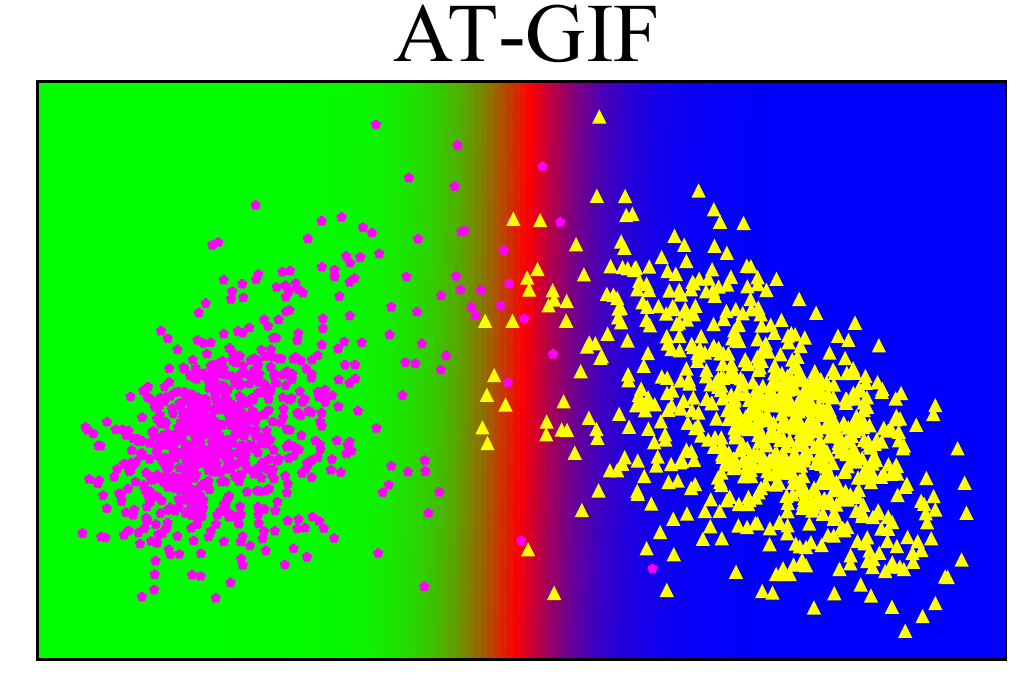}
	}
	\\
% 	\hspace{14.52mm}
    %\hspace{80mm}
    %\hspace*{\fill}%
% 	\hspace{8mm}
    \subfigure{
		\includegraphics[width=0.8\linewidth]{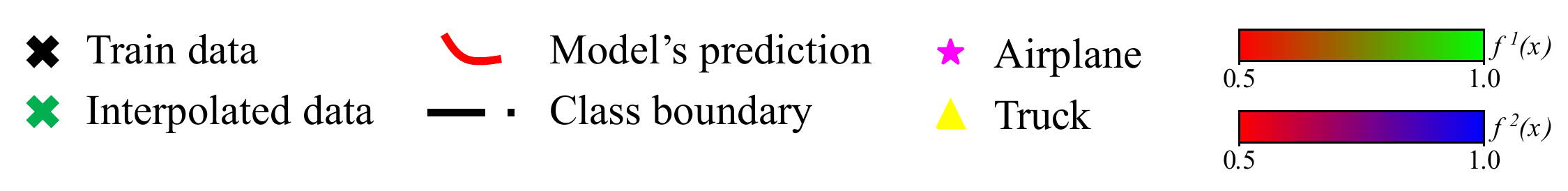}
	}
% 	\subfigure{
% 		\includegraphics[width=0.35\linewidth]{figures/decision_boundary_legend.pdf}
% 	}
  %\vspace{-3mm}
  \caption{Comparisons between AT, AT-mixup, and AT-GIF. \textbf{The three upper panels} illustrate AT, AT-mixup, and AT-GIF, respectively. The horizontal axis is the input feature, and the vertical axis is the prediction of data. The data in blue shade are predicted as the first label, and the data outside the blue shade are predicted as the second label. 
  \textbf{The three lower panels} show the experimental results of AT, AT-mixup, and AT-GIF, respectively. The data are randomly selected from the CIFAR-10 dataset.
  %We randomly select 1600 data from 2 random classes (800 data for each class) from the CIFAR-10 dataset. We embed the data to 2-dimension with PCA~\citep{abdi2010principal}. Then, we train these 2D data on a small CNN. 
  The detailed settings are in Appendix \ref{app_sec:fig12}. 
  Red-green shading represents $f^1(x)$ (the prediction confidence for ``airplane"), and red-blue shading represents $f^2(x)$ (the prediction confidence for ``truck"). %: A darker shade represents a larger prediction confidence. 
  \textbf{The two left panels} indicate that the standard AT has a sharp transition between the two classes. %The upper left panel demonstrates in AT, the adversarial variants of the two test data can be easily misclassified, due to a low ratio of . 
  \textbf{The two middle panels} demonstrate that AT-mixup leads to the model's linear behavior between classes: the prediction confidence changes smoothly. 
  \textbf{The two right panels}
   suggest that AT-GIF can obtain more attackable data by interpolation (see Figure \ref{fig:att_acc}) and still keeps a sharper transition between the two classes: the prediction confidence changes abruptly. Besides, in Appendix \ref{app_sec:aux_exp}, we conduct an auxiliary experiment to further compare the differences between AT, AT-mixup, and AT-GIF on ResNet-18~\citep{he2016deep}. We visualize the linear behavior of AT-mixup, and show that AT-GIF can alleviate the linear behavior.
}
  \label{fig:linear_behavior}
  %\vspace{-4mm}
\end{figure*}

\subsubsection{Unequal treatment of data in AT}
\label{sec:at_unequal}
Recent studies~\citep{ding2020mma,wang2020improving_MART,cheng2020cat,zhang2021geometry} showed that data do not contribute equally during adversarial training.
Max-margin adversarial training (MMA)~\citep{ding2020mma} demonstrated that adversarial data close to the model's decision boundary are more important to the robust model, while adversarial data far from the decision boundary are less important.
Misclassification aware adversarial training (MART)~\citep{wang2020improving_MART} differentiated the misclassified and correctly classified data during AT.
Customized adversarial training (CAT)~\citep{cheng2020cat} adaptively customized the perturbation bound and the corresponding label for each training data according to their distance to the decision boundary.
Geometry-aware instance-reweighted adversarial training (GAIRAT)~\citep{zhang2021geometry} paid more attention to data close to the decision boundary.
Specifically, GAIRAT split all the data into two categories---\textit{attackable data} and \textit{guarded data}.
Data close to the decision boundary are called attackable data, while guarded data are far from the decision boundary. 
Formally, a sample $(x,y)$ is said to be attackable, if
\begin{equation}\label{eq:attackable}
\begin{split}
\argmax_c f^c_\theta(\tilde{x})\neq\argmax_c y^c,
\end{split}
\end{equation}
where $\tilde{x}$ is the adversarial variant of natural data $x$.
On the other hand, a sample $(x,y)$ is said to be guarded, if
\begin{equation}\label{eq:guarded}
\begin{split}
\argmax_c f^c_\theta(\tilde{x})=\argmax_c y^c.
\end{split}
\end{equation}
%Appendix A shows an illustration of attackable data and guarded data.
%Data in the shaded region are attackable data, and their adversarial variants can be easily misclassified. Data outside the shaded region are guarded data. Their adversarial variants are hard to be misclassified. 
The adversarial variants of attackable data can be easily misclassified, while guarded data's adversarial variants are hard to be misclassified.
GAIRAT claimed that attackable data help to improve adversarial training, while the guarded data are less important.

We propose a guided interpolation framework (GIF), which utilizes the meta information of the previous epoch to guide the data's interpolation. The GIF can obtain a high ratio of attackable data to enhance AT. Compared to the widely used vanilla mixup~\citep{zhang2018mixup}, the GIF can obtain more attackable data by interpolation and mitigate the linear behavior of the vanilla mixup (see Figure \ref{fig:att_acc}, \ref{fig:linear_behavior} and Section \ref{sec:GIF_detail} for details), which is beneficial to AT. Furthermore, the GIF can be incorporated into existing AT methods (e.g., AT~\citep{Madry_adversarial_training}, TRADES~\citep{zhang2019theoretically}, GAIRAT~\citep{zhang2021geometry}, and FastAT~\citep{wong2020fast_zico_kolter}) and improve their robustness (see Section \ref{sec:exp_robustness} for details).

\begin{algorithm}[!t]
    %\setstretch{2.35}
    \caption{Adversarial training with guided interpolation framework (AT-GIF)}
    \label{alg:GIF}
    %\hspace*{\algorithmicindent} 
    \textbf{Input:} model $f_\theta$, training dataset $S$, learning rate $\eta$, number of epochs $T$, original batch size $m$, interpolation batch size $m'$, number of batches $M$, perturbation bound $\epsilon$, step size $\alpha$, PGD step number $K$ \\
    \textbf{Output:} adversarially robust network $f_\theta$
    \begin{algorithmic} % The number tells where the line numbering should start
        \State $A_0\gets S$
        %\Comment{Initialize attackable set.}
        %\State $\mathcal{G}_0\gets\emptyset$
        \For{$\text{epoch } t= 1, 2, ..., T$}
            \State $A_t=\emptyset$
            \State Compute interpolation set $\overline{S}_t$ by $A_{t-1}$ and Eq. (\ref{eq:gif})
            % \For{$i=1,..., n$} \Comment{Generate interpolation set.}
            %     \State Randomly sample two data $(x_{i,1},y_{i,1}), (x_{i,2},y_{i,2})$ from $A_{t-1}$
            %     \State $\overline{x}_i\gets 0.5*x_{i,1}+0.5*x_{i,2}$
            %     \Comment{Interpolate data according to Eq. (\ref{eq:gif}).}
            %     \State $\overline{y}_i\gets 0.5*y_{i,1}+0.5*y_{i,2}$
            %     \State $\overline{S}_t\gets\overline{S}_t\cup{(\overline{x}_i,\overline{y}_i)}$
            % \EndFor
            %\State $\overline{S}_t\gets$IntSetGen$(A_{t-1},n)$ 
            \For{$\text{mini-batch }=1,..., M$}
                \State Randomly sample $\{(x_i,y_i)\}_{i=1}^{m}$ from $S$
                \For{$i=1, \ldots, m$} {\textbf{in parallel}} %\Comment{Train with original data and interpolated data.}
                    \State Generate adversarial data $\widetilde{x}_i$ of $x_i$ by Eq. (\ref{eq:pgd})
                    % \State $\widetilde{x}_i\gets x_i$
                    % \For{$k=1, \ldots, K$}
                    %     \State $\widetilde{x}_i\gets\Pi_{\mathcal{B}_\epsilon(x_i)}\left(\alpha \text{sign}\left(\nabla_{\widetilde{x}_i}\ell(f_\theta(\widetilde{x}_i),y_i)\right)+\widetilde{x}_i\right)$
                    %     \Comment{Find adversarial data according to Eq. (\ref{eq:at_adv_x}).}
                    % \EndFor
                    \If{ $\argmax_c f^c_\theta(\tilde{x}_i)\neq\argmax_c y_i^c$}
                    %\Comment{Test the original data according to Eq. (\ref{eq:attackable}).}
                    \State $A_t\gets A_t\cup\{(x_i,y_i)\}$
                    %\Comment{Update the attackable set.}
                    \EndIf
                \EndFor
                %\State Sample $\{(\overline{x}_i,\overline{y}_i)\}_{i=m+1}^{m+m'}$ from ${\overline{S}_t}$
                \State Randomly sample $\{(x_i,y_i)\}_{i=m+1}^{m+m'}$ from ${\overline{S}_t}$
                \For{$i=m+1,...,m+m'$} {\textbf{in parallel}}
                    \State Generate adversarial data $\widetilde{x}_i$ of $x_i$ by Eq. (\ref{eq:pgd})
                % \Comment{Train with interpolated data.}
                %     \State $\widetilde{x}_i\gets\overline{x}_i$
                %     \For{$k=1,...,K$}
                %         \State $\widetilde{x}_i\gets\Pi_{\mathcal{B}_\epsilon(\overline{x}_i)}\left(\alpha \text{sign}\left(\nabla_{\widetilde{x}_i}\ell(f_\theta(\widetilde{x}_i),\overline{y}_i)\right)+\widetilde{x}_i\right)$
                %         \Comment{Find adversarial data according to Eq. (\ref{eq:at_adv_x}).}
                %     \EndFor
                \EndFor
                \State $\theta\gets\theta-\eta\cdot\frac{1}{m+m'}\sum_{i=1}^{m+m'} \nabla_\theta\ell(f_\theta(\widetilde{x}_i),y_i)$
                %\Comment{Update model parameters.}
            \EndFor
        \EndFor
    \end{algorithmic}
\end{algorithm}

\subsection{Mixup for Robustness}
\label{sec:Review_mixup}
In standard training (ST), mixup has been widely used to improve the generalization~\citep{zhang2018mixup,thulasidasan2019mixup,NEURIPS2019_1cd138d0,Berthelot2020ReMixMatch,kim2021comixup,zhang2021how}.
Mixup augments the original dataset with interpolated data as follows.
\begin{equation}\label{eq:mixup}
\begin{split}
\overline{x}&=\lambda x_i+(1-\lambda)x_j, \\
\overline{y}&=\lambda y_i+(1-\lambda)y_j,
\end{split}
\end{equation}
where $(\overline{x},\overline{y})$ is the interpolated data, $(x_i,y_i)$ and $(x_j,y_j)$ are two randomly selected data from the original dataset $S$, and $\lambda$ is the interpolation weight. 
~\citet{zhang2021how} showed that mixup (in ST) can improve the robustness against fast gradient sign attack (FGSM)~\citep{goodfellow2014explaining}. 
%Nevertheless, the robustness of mixup (in ST) against stronger adaptive attacks (e.g., PGD~\citep{Madry_adversarial_training} and CW attack~\citep{Carlini017_CW}) is not discussed in \citep{zhang2021how}.
Nevertheless, mixup (in ST) fails to defend stronger adaptive attacks (e.g., PGD~\citep{Madry_adversarial_training} and CW attack~\citep{Carlini017_CW})~\citep{zhang2018mixup}. 
%To solve this problem, we propose the GIF for AT, which aims to defend our neural networks from such strong adaptive attacks.

To defend against adaptive attacks, several mixup-based adversarial training techniques have been proposed~\citep{lamb2019interpolated,laugros2020addressing}.
Interpolated adversarial training (IAT)~\citep{lamb2019interpolated} trains on interpolations of adversarial data along with interpolations of natural data. 
%combines 
%adopted vanilla mixup in adversarial training to improve the adversarial robustness. Note that IAT is different from AT-mixup (shown in Figure \ref{fig:att_acc} and \ref{fig:linear_behavior}).
Mixup with targeted labeling adversarial training
(M-TLAT)~\citep{laugros2020addressing} combines vanilla mixup with targeted labeling to enhance AT. 
However, these methods behave linearly when employing vanilla mixup in AT, and such linear behaviors will damage the adversarial robustness (see Section \ref{sec:Motivation} for details). 
To alleviate this issue, we propose the GIF for AT, which aims to obtain a high ratio of attackable data by interpolation and mitigate the linear behavior (see Section \ref{sec:GIF_detail} for details).

% By interpolating the original data, mixup~\citep{zhang2018mixup} can obtain additional data information and enhance traditional supervised learning.\JF{delete this sentence.}
% \JF{Instead, you need to state this work \citep{zhang2021how}: Notably, vanilla mixup can enhance robustness of ST, but can only agaist FGSM attack but not adaptive attack. Our GIF works for AT, which aim to defend adaptive attack.}

\citet{pang2019mixup} proposed mixup inference (MI) to exploit the linear behaviors of the mixup-based models in the inference phase, while our work concentrates on investigating data interpolation in the training phase.
%However, linearity only exist in methods based on vanilla mixup, which might not be suitable for other methods.

% To mitigate the downtrend of attackable data and overfitting issue, mixup-based adversarial training techniques are proposed~\citep{lamb2019interpolated,lee2020adversarial}.
% Interpolated adversarial training (IAT)~\citep{lamb2019interpolated} adopted vanilla mixup in adversarial training to improve adversarial robustness.
% Adversarial vertex mixup (AVmixup)~\citep{lee2020adversarial} combined mixup with label smoothing to improve robust generalization.
% By utilizing mixup in adversarial training, these methods can enrich the training data, thereby increasing the number of attackable data.

%Although these mixup-based methods can improve robustness, mixup is designed for ST, which might bring hardness in AT. Furthermore, \citet{zhang2021how} only demonstrates vanilla mixup can improve robustness against one-step adversarial attack, i.e. FGSM~\citep{goodfellow2014explaining}. Nevertheless, in adversarial training, multi-step PGD is a widely used stronger attack, which is not mentioned in \citep{zhang2021how}. 

\section{Guided Interpolation Framework (GIF)}
\label{sec:GIF}
In this section, we present our motivation and introduce our proposed guided interpolation framework (GIF).

%\paragraph{Motivation of GIF.}
\subsection{Motivation of GIF}
\label{sec:Motivation}
\paragraph{Enhancing adversarial robustness further requires abundant attackable data.} As discussed in Section \ref{sec:at_unequal}, data contribute unequally during AT, and more attackable data can better benefit the robustness enhancement. The left panel of Figure \ref{fig:att_acc} indicates both vanilla mixup (orange dotted line)~\citep{zhang2018mixup} and the GIF (blue dotted line) can introduce a higher ratio of attackable data (compared to standard AT). the right panel of Figure \ref{fig:att_acc} shows AT-mixup and AT-GIF can indeed improve adversarial robustness. 
This suggests that utilizing mixup or GIF is helpful to AT.

\paragraph{Adversarial robustness requires the local invariance.} \citet{goodfellow2014explaining},  ~\citet{papernot2016towards}, and~\citet{zhang2021geometry} demonstrated that in AT, the model predictions are supposed to be locally invariant to the inputs' neighborhood.
~\citet{ding2020mma} proposed to maximize the distances from the inputs to the model's decision boundary, which encouraged the invariant predictions in the inputs' neighbourhood.
% minimize the maximum loss within the inputs' neighborhood\JF{rephrase it to make it clearer}.
The above studies all show that a well-trained adversarial model should be locally invariant. This infers the model's linear behavior is never AT's desideratum.

However, directly employing vanilla mixup in AT will lead to the linear behavior. 
%In Figure \ref{fig:linear_behavior}, we compare AT-mixup (two middle panels) with standard AT (two left panels). 
Figure \ref{fig:linear_behavior} shows that the prediction of AT-mixup (two middle panels) moves linearly from one class to the other, while standard AT (two left panels) has a sharp transition between the two classes.
%Figure \ref{fig:toy_illustration} also illustrates this fact, which AT-mixup has a more linear prediction than AT.
%This linear behavior of AT-mixup is not in favor of AT.
Although AT-mixup can introduce more attackable data to improve adversarial robustness, it inevitably involves undesirable linear behavior in AT.
%, which will undermine the adversarial robustness. Thus, naively introducing vanilla mixup in AT might have little effect in robustness enhancement. [although more attackable data, linear behavior]

The above observation motivates us to obtain more attackable data by interpolation meanwhile without sacrificing much the local invariance.

\subsection{Realization of GIF}
\label{sec:GIF_detail}
In mixup, each interpolated data is computed by two different data (Eq. (\ref{eq:mixup})), which results in a soft interpolated label $\overline{y}\in[0,1]^C$.
Let $(\overline{x},\overline{y})$ be an interpolated data computed by $(x_i,y_i)$ and $(x_j,y_j)$ (with Eq. (\ref{eq:mixup})), $\tilde{x}$ be the adversarial variant of $\overline{x}$ computed with Eq. (\ref{eq:pgd}). 
We say the interpolated data $(\overline{x},\overline{y})$ is attackable, if
\begin{equation}\label{eq:gif_attackable}
\begin{split}
\argmax_c f^c_\theta(\tilde{x})\neq\argmax_c y_i^c \land
\argmax_c f^c_\theta(\tilde{x})\neq\argmax_c y_j^c,
\end{split}
\end{equation}
where $\land$ is ``And" operation. Otherwise, we say it is guarded, if
\begin{equation}\label{eq:gif_guarded}
\begin{split}
\argmax_c f^c_\theta(\tilde{x})=\argmax_c y_i^c \lor
\argmax_c f^c_\theta(\tilde{x})=\argmax_c y_j^c,
\end{split}
\end{equation}
where $\lor$ is ``Or" operation.
% Let $(\overline{x},\overline{y})$ be an interpolated data, $\tilde{x}$ be the adversarial variant of $\overline{x}$, and $c=\argmax_c f^c_\theta(\tilde{x})$ be the predicted label of $\tilde{x}$.
% The interpolated data $(\overline{x},\overline{y})$ is attackable, if 
% \begin{equation}\label{eq:gif_attackable}
% \begin{split}
% \overline{y}^c=0.
% \end{split}
% \end{equation}
% The interpolated data $(\overline{x},\overline{y})$ is guarded, if 
% \begin{equation}\label{eq:gif_guard1}
% \begin{split}
% \overline{y}^c\neq 0.
% \end{split}
% \end{equation}

According to Eq. (\ref{eq:mixup}), there are two core factors in data interpolation: 1) selecting data for interpolation and 2) setting the interpolation weight $\lambda$.

The first part is selecting data for interpolation. 
Vanilla mixup randomly picks data for interpolation. The attackable ratio of interpolated data also drops rapidly as adversarial training progresses (see the orange dotted line in the left panel of Figure \ref{fig:att_acc}).
To alleviate this issue, our GIF only chooses attackable data (see Eq.\eqref{eq:attackable}) for interpolation; thus, the interpolated data are more likely to be attackable (see the blue dotted line in the left panel of Figure \ref{fig:att_acc}). 

The second part is setting the interpolation weight $\lambda$. In vanilla mixup, $\lambda$ is sampled randomly from a uniform (or beta) distribution, which is not suitable in adversarial training. A uniformly sampled $\lambda$ will encourage the model to behave linearly between classes, which is not in favor of AT. 
The two middle panels of Figure \ref{fig:linear_behavior} show that AT-mixup has a linear transition, which is not favorable to AT.
Instead of random sampling from a uniform distribution, we argue to fix $\lambda=0.5$. 
%First, by fixing $\lambda=0.5$, the interpolated data is likely to be close to the decision boundary, which aids in fine-tuning the decision boundary. \CC{We conduct experiments with different $\lambda$ in Appendix XXX}\JF{Do not forget this.} to verify this fact. Hence, we have a higher chance to obtain attackable data.
By fixing $\lambda=0.5$, we can mitigate the linear behavior between classes (see AT-GIF in Figure \ref{fig:linear_behavior}). 
Thus, fixing $\lambda=0.5$ will benefit AT.
We also discuss the effect of different interpolation weight $\lambda$ in Section \ref{sec:ablation}.

%Motivated by the above observation, we interploate data by Alg. \ref{alg:interpolate}.
In summary, we interpolate data as follows:
\begin{equation}\label{eq:gif}
\begin{split}
\overline{x}=0.5 x_i+0.5 x_j \\
\overline{y}=0.5 y_i+0.5 y_j,
\end{split}
\end{equation}
where $(\overline{x},\overline{y})$ is the interpolated data, $(x_i,y_i)$ and $(x_j,y_j)$ are two random attackable data satisfying Eq. (\ref{eq:attackable}). 
%The interpolation procedure is demonstrated in Algorithm \ref{alg:interpolate}\footnote{JF: I guess you do not need Algorithm 2. }.
In the left panel of Figure \ref{fig:att_acc}, the attackable ratio of data interpolated by the GIF (blue dotted line) is stabled at around 90\%, which indicates our interpolation strategy can indeed increase the ratio of attackable data.
Besides, the two right panels of Figure \ref{fig:linear_behavior} also demonstrate that AT-GIF (with $\lambda=0.5$) behaves sharply between classes. Therefore, our GIF is beneficial to the enhancement of the adversarial robustness. 

The training of AT-GIF is shown in Algorithm \ref{alg:GIF}.
First, at the beginning of each epoch, we obtain the interpolation set from the previous attackable set (computed in the previous epoch).
%Second, in each mini-batch, we sample two batches of data: one from the original dataset and the other from the interpolation set.
Second, we generate the adversarial variants of data from the original dataset along with the interpolation set for updating the model.
%these two batches of data for updating the model.
Third, we filter out the attackable data from the original dataset and put them into the current attackable set $A$. This attackable set $A$ will be used to obtain the interpolation set that guides the next-epoch training. 
Last, we compute the loss of the adversarial variants and update the parameters $\theta$ with the gradient of the loss.

In Algorithm \ref{alg:GIF}, a burn-in period may be introduced. During the initial period of the training epochs, instead of generating data with the GIF, we utilize the original data without the need for the interpolated data for the AT. This is because, during the initial period, standard AT has a high ratio of attackable data which is enough for the enhancement of the adversarial robustness. We also discuss the effect of the burn-in period in Section \ref{sec:ablation}.

Figure \ref{fig:att_acc} and \ref{fig:linear_behavior} show the differences between standard AT, AT-mixup (AT with vanilla mixup), and AT-GIF (AT with GIF). 
The left panel of Figure \ref{fig:att_acc} demonstrates that the GIF can obtain nearly 90\% of attackable data by interpolation (blue dotted line), which is much higher than the interpolated data by vanilla mixup (orange dotted line) and the original data (red line). 
The right panel of Figure \ref{fig:att_acc} manifests that our AT-GIF can indeed improve robustness with a higher ratio of attackable data.
Figure \ref{fig:linear_behavior} indicates that AT-mixup (two middle panels) leads to the robust model that behaves linearly, while the robust models trained by AT (two left panels) and AT-GIF (two right panels) have a sharper transition between different classes; inside each class, the model's predictions are almost invariant. 
Therefore, AT-GIF can obtain more attackable data by the interpolation and meanwhile keep the local invariance. 
As a result, AT-GIF can lead to a more robust model against the adversarial attacks.
% Figure \ref{fig:linear_behavior} shows the differences between standard AT, adversarial training with vanilla mixup (AT-mixup), and adversarial training with GIF (AT-GIF). 
% Standard AT (two upper panels) has few attackable data and can be easily attacked.
% %Standard AT has few attackable data. Thus, the decision boundary is not stable\footnote{JF: what does this mean?}, and can be easily attacked. 
% AT-mixup (two middle panels) behaves linearly, which is not suitable in AT. 
% AT-GIF (two lower panels) has a sharp transition between different classes. Moreover, AT-GIF has a higher ratio of attackable data (see the blue solid line in the left panel of Figure \ref{fig:att_acc})
% , which can benefit AT. Thus, AT-GIF is more robust to adversarial attacks.

\begin{table*}[!t]
\scriptsize %\addtolength{\tabcolsep}{-5pt}
\centering
\normalsize
\caption{Test accuracy of ResNet-18 on the CIFAR-10 dataset.}
\label{tbl:resnet18_acc}
%\resizebox{1.99\columnwidth}{!}
{
\begin{tabular}{c|cccc|cccc}
\hline
\multirow{2}*{Defense} & \multicolumn{4}{c|}{Best} &\multicolumn{4}{c}{Last} \\
\cline{2-9}
&Natural & PGD-20 &   CW & AA &  Natural &  PGD-20 &  CW & AA  \\
\hline
AT~\citep{Madry_adversarial_training} &  81.72 & 51.05  & 49.86 &  47.25 & 84.77 &   45.47 &  45.90 &  43.28  \\
%\hline
IAT~\citep{lamb2019interpolated} & \textbf{90.61} & 41.11 & 41.93 & 36.43 & \textbf{90.53} & 39.50 & 40.43 & 35.10 \\
%\hline
$KD_{std\&adv} + SWA$~\citep{chen2021robust} & 85.16 & 46.85 & 46.73 & 44.74 & 85.63 & 45.45 & 46.12 & 43.70 \\
%\hline
AT-GIF (Ours) & 81.59 & \textbf{53.57}  & \textbf{50.07}   & \textbf{47.91} & 83.15 & \textbf{52.06} &  \textbf{48.21}  & \textbf{45.59}   \\
\hline
\end{tabular}
}
%\vspace{-15pt}
\end{table*}

\begin{table*}[!t]
\scriptsize %\addtolength{\tabcolsep}{-5pt}
\centering
\normalsize
\caption{Test accuracy of WRN-32-10 on the CIFAR-10 dataset.}
\label{tbl:wrn_acc}
%\resizebox{1.99\columnwidth}{!}
{
\begin{tabular}{c|cccc|cccc}
\hline
\multirow{2}*{Defense} & \multicolumn{4}{c|}{Best} &\multicolumn{4}{c}{Last} \\
\cline{2-9}
&Natural &PGD-20 &  CW &  AA & Natural &  PGD-20 &  CW &  AA  \\
\hline
AT~\citep{Madry_adversarial_training} & 86.73 & 53.66  & 53.66 & 50.96 & 86.88 & 47.94 &  48.12 &  45.82 \\
%\hline
IAT~\citep{lamb2019interpolated} & \textbf{95.79} & 50.85 & 49.53 & 11.51 & \textbf{96.55} & 47.37 & 48.50 & 3.87 \\
%\hline
$KD_{std\&adv} + SWA$~\citep{chen2021robust} & 84.46 & 52.73 & 52.53 & 50.85 & 86.53 & 50.34 & \textbf{50.29} & \textbf{48.07} \\
%\hline
AT-GIF (Ours) & 86.49 & \textbf{55.92} &  \textbf{53.99} &  \textbf{51.92} & 86.14 & \textbf{51.42} & 49.17 &  46.70  \\
\hline %\cellcolor{gray!30}
\end{tabular}
}
%\vspace{-15pt}
\end{table*}

%and employing GIF requires addition computation cost.

%GIF has two main differences with standard AT. First, GIF need to testing every data in order to generate the attackable set. Second, GIF need to interpolate data with the attackable set generated 

%Besides attackable data, XXXX\footnote{other data choosing methods} add a figure here.

%\subsection{Realization of GIF}

%GIF prevents overfitting

%$\lambda$: add a illustration here to explain why we want 0.5. add a discussion here, to show that hard label does not work. hard label cannot work well:  If we require the label to be the same after mixup, we may extend the boundary of a class *too much*; the situation gets worse if we further significantly smooth the eps-ball around hard-label mixup data. This is similar to the case why people keep the same eps and change FGSM to PGD rather than using bigger eps for FGSM.

%$\lambda$: add a illustration here to explain why we want 0.5.
%In GIF, we set the interpolation weight $\lambda$ to be $0.5$. [more details]
%1. 0.5 have more attackable data.
%2. the data make the decision boundary smooth (by illustration).

%real reason (might be?): the model should be linear/ smooth around decision boundary, sharp in other place.
%choice

\section{Experiments}
\label{sec:Exp}
%\footnote{JF: this section, you need divide it into two sections. Section 4.1 Robustness Evaluation. Section 4.2 Ablation Study.}
In this section, we empirically justify the efficacy of our proposed GIF on CIFAR-10~\citep{krizhevsky2009learning} and SVHN~\citep{netzer2011reading_SVHN} datasets. The detailed descriptions of the two datasets are in Appendix \ref{app_sec:dataset}. All models are trained with ResNet-18~\citep{he2016deep} or Wide ResNet (WRN)~\citep{zagoruyko2016WRN}. 
All experiments are run on the same machine with Intel Xeon Gold 5218 CPU, 250GB RAM, and six NVIDIA Tesla V100 SXM2 GPU. All methods are implemented with PyTorch 1.7.0~\citep{paszke2019pytorch}.

%\subsection{Settings}
\paragraph{Setup.}
We first introduce the settings for the CIFAR-10 dataset.
In our experiments, we consider $||\tilde{x}-x||_{\infty}<\epsilon$ with the same $\epsilon$ in both training and evaluations.
All images of CIFAR-10 are normalized into $[0,1]$.
We set the original batch size $m=64$; the interpolation batch size $m'=64$. 
For generating the most adversarial data to update the model, the perturbation bound $\epsilon=8/255$; PGD step number $K=10$; and PGD step size $\alpha=2/255$, which keeps the same as~\citep{rice2020overfitting}.
We train the model using SGD with $0.9$ momentum for 60 epochs with the initial learning rate of $0.1$ divided by 10 at Epoch 30 and 45, respectively. The weight decay is $5e-4$. 
For evaluations, we report standard test accuracy for natural test data and robust test accuracy for adversarial test data. The adversarial test data are generated by PGD-20 attack (PGD attack with 20 steps)~\citep{Madry_adversarial_training}, CW attack (with 30 steps)~\citep{Carlini017_CW}, and Auto attack (AA)~\citep{croce2020reliable} with the same perturbation bound $\epsilon=8/255$. The step size $\alpha$ for PGD-20 attack and CW attack is $2/255$. 

In SVHN, we set the perturbation bound $\epsilon=4/255$, PGD step size $\alpha=1/255$, and initial learning rate $\eta=0.01$. For evaluations, the adversarial test data are generated by PGD-20 attack, CW attack, and Auto attack (AA) with the same perturbation bound $\epsilon=4/255$. The step size $\alpha$ for PGD-20 attack and CW attack is $1/255$. Other settings of SVHN are the same as CIFAR-10.

%\subsection{Baselines}
We compare our AT-GIF with the following methods:
standard adversarial training (AT)~\citep{Madry_adversarial_training}, IAT~\citep{lamb2019interpolated},
$KD_{std\&adv} + SWA$~\citep{chen2021robust}. 
We also modify TRADES~\citep{zhang2019theoretically}, GAIRAT~\citep{zhang2021geometry}, and FastAT~\citep{wong2020fast_zico_kolter} to their GIF versions, i.e., TRADES-GIF, GAIRAT-GIF, and FastAT-GIF, and compare the performance of these methods. The algorithm of TRADES-GIF, GAIRAT-GIF, and FastAT-GIF are shown in Appendix \ref{app_sec:alg}. 

For fair comparisons, instead of sampling interpolated data from the interpolation set $\overline{S}$ in GIF-based methods, all baseline methods set the original batch size $m=128$ and the interpolation batch size $m'=0$, i.e., we make sure that all methods train on the same amount of data. 
All GIF-based methods (e.g., AT-GIF, TRADES-GIF, GAIRAT-GIF, and FastAT-GIF) involve a burn-in period except the models trained in Figure \ref{fig:att_acc} and Figure \ref{fig:linear_behavior}. The detailed settings of all these methods are in Appendix \ref{app_sec:baseline}.
%In the first 30 epochs, these methods train with their naive version (e.g., AT, TRADES, and GAIRAT). After Epoch 30, they start to interpolate data with GIF and train the model with both original data and interpolated data. 

%In Figure \ref{fig:att_acc}, all methods are trained with ResNet-18. In Figure \ref{fig:mixup_bound}, the methods are trained on a small neural network with four fully connected layers (size (10,10,5,2)). To better illustrate our GIF, the models trained in Figure \ref{fig:att_acc} and \ref{fig:mixup_bound} do not involve the burn-in period. %Other settings are the 

% \begin{table*}[!t]
% \scriptsize %\addtolength{\tabcolsep}{-5pt}
% \centering
% \normalsize
% \caption{Test accuracy of FGSM and FGSM-GIF on CIFAR-10 dataset.}
% \label{tbl:fgsm}
% %\resizebox{1.99\columnwidth}{!}
% {
% \begin{tabular}{c|cccc|cccc}
% \hline
% \multirow{2}*{Defense} & \multicolumn{4}{c|}{ResNet-18} &\multicolumn{4}{c}{WRN-32-10} \\
% \cline{2-9}
% & Natural &PGD-20 &  CW &  AA & Natural &  PGD-20 &  CW &  AA  \\
% \hline
% FGSM & \textbf{86.33} & 43.81 &  45.12 &  41.03 & 88.38 & 46.14  & 47.31  & \textbf{43.70} \\
% %\hline
% FGSM-GIF & 85.45 & \textbf{47.49} & \textbf{45.88} & \textbf{42.88} & \textbf{88.83} & \textbf{48.42} & \textbf{48.26} &  41.91 \\
% \hline
% \end{tabular}
% }
% %\vspace{-15pt}
% \end{table*}

\subsection{Robustness Evaluation}
\label{sec:exp_robustness}
%In this section, we evaluate the performance of our GIF and show that our GIF can enhance the adversarial robustness.
In this section, we first evaluate the robustness of AT-GIF on the CIFAR-10 dataset. Then, we show that the GIF can be easily incorporated into modern adversarial training methods (e.g., TRADES, GAIRAT, and FastAT). Last, we demonstrate that the GIF can also improve robustness on the SVHN dataset.

\paragraph{Robustness evaluation on CIFAR-10.}
We first show the efficiency of the GIF on CIFAR-10.
%To achieve a better performance, we adopt the default setting of IAT~\citep{lamb2019interpolated}. They are trained for 100 epoches, and the learning rate is divided by 10 at Epoch 50 and 75 respectively.
We compare AT-GIF with standard AT, IAT, and $KD_{std\&adv} + SWA$. 

In Table \ref{tbl:resnet18_acc}, we report the test accuracy of these models trained with ResNet-18.
To keep the same setting as~\citep{Madry_adversarial_training}, we also report the performance of WRN-32-10 in Table \ref{tbl:wrn_acc}.
AT-GIF achieves the best adversarial robustness. For example, AT-GIF improves the best robust accuracy of AT by 2.52\% on ResNet-18 under PGD-20 attack. The results indicate that AT suffers from overfitting, and has a large robust accuracy drop in the last epoch. AT-GIF alleviates the overfitting issue by interpolating more attackable data. 
The results also demonstrate that AT-GIF outperforms vanilla mixup-based methods (e.g., IAT), which further verifies the design of the GIF is more suitable (than vanilla mixup) in AT. 
$KD_{std\&adv} + SWA$ mitigates the overfitting problem, but can hardly improve the robustness. AT-GIF can alleviate the overfitting issue and improve robustness at the same time. 

\begin{table*}[!t]
\scriptsize %\addtolength{\tabcolsep}{-5pt}
\centering
\normalsize
\caption{Test accuracy of GIF incorporating with other AT methods on the CIFAR-10 dataset.}
\label{tbl:gif+other}
%\resizebox{1.99\columnwidth}{!}
{
\begin{tabular}{c|c|cccc|cccc}
\hline
\multirow{2}*{Defense} & \multirow{2}*{Network} & \multicolumn{4}{c|}{Best} &\multicolumn{4}{c}{Last} \\
\cline{3-10}
& & Natural & PGD-20 &  CW &  AA & Natural & PGD-20 & CW & AA \\
\hline
TRADES & \multirow{2}*{WRN-34-10} & \textbf{84.92} & 55.33  & 53.81  & 52.54 &  \textbf{85.10} & 53.16 & 51.12 & 50.09 \\
TRADES-GIF & & 83.64 & \textbf{57.45} & \textbf{54.14} & \textbf{53.37} & 84.25 & \textbf{56.91} & \textbf{53.91} & \textbf{53.08}  \\
\hline
GAIRAT &\multirow{2}*{WRN-32-10} & \textbf{86.31} & 56.69 &  44.63 &  42.26 & \textbf{85.43}  & 52.50 & 43.71 & 41.14 \\
GAIRAT-GIF &  & 84.59 &\textbf{64.97}  & \textbf{46.71} & \textbf{44.84} & 82.18 & \textbf{55.11} & \textbf{45.65} & \textbf{43.24}\\
\hline 
FastAT & \multirow{2}*{WRN-32-10} & 88.38 & 46.14  & 47.31  & \textbf{43.70}  & 89.56  & 6.79 & 6.01 & 0.03 \\
FastAT-GIF &  & \textbf{88.83} & \textbf{48.42} & \textbf{48.26}  & 41.91  & 89.93  & 0.39  & 0.34 & 0.00 \\
\hline %\cellcolor{gray!30}
\end{tabular}
}
%\vspace{-15pt}
\end{table*}

\paragraph{Robustness evaluation of GIF incorporating with other AT methods.}
%\label{sec:other_methods}
To show the GIF is a compatible approach, we compare TRADES, GAIRAT, and FastAT with their GIF versions, i.e., TRADES-GIF, GAIRAT-GIF, and FastAT-GIF.
%we modify TRADES, GAIRAT, and FastAT to their GIF versions, i.e., TRADES-GIF, GAIRAT-GIF, and FastAT-GIF, and compare the performance of these methods.

Table \ref{tbl:gif+other} demonstrates the results of these methods trained with Wide ResNet on the CIFAR-10 dataset. To make fair comparisons, TRADES and TRADES-GIF are trained with WRN-34-10~\citep{zhang2019theoretically}, and other methods are trained with WRN-32-10~\citep{zhang2021geometry,wong2020fast_zico_kolter}. We also report the results of these models trained with ResNet-18 on the CIFAR-10 dataset in Appendix \ref{app_sec:gif+others}.
The GIF versions (TRADES-GIF, GAIRAT-GIF, and FastAT-GIF) have a better performance than their original versions (TRADES, GAIRAT, and FastAT). For instance, GAIRAT-GIF increases the robust test accuracy of GAIRAT by 8.28\% on WRN-32-10 under PGD-20 attack.
The failure of FastAT and FastAT-GIF in the last epoch is due to catastrophic overfitting~\citep{wong2020fast_zico_kolter}. 

% \paragraph{Performance evaluation of FGSM.}
% %We demonstrate the results of FGSM on CIFAR-10 dataset.
% We train the model with FGSM on  CIFAR-10 dataset. We set step number $K=1$, step size $\alpha=8/255$, and the perturbation bound $\epsilon=8/255$. For evaluation, the adversarial test data are generated by FGSM with step number $K=1$, step size $\alpha=8/255$, and perturbation bound $\epsilon=8/255$.

% Table \ref{tbl:fgsm} shows the results of FGSM and its GIF version, i.e., FGSM-GIF. FGSM-GIF significantly boosts the performance of FGSM.

\paragraph{Robustness evaluation on SVHN.}
We report the performance of AT-GIF compared with AT on the SVHN dataset in Table~\ref{tbl:svhn}.
% We conduct the experiments on SVHN dataset.
% Table \ref{tbl:svhn} demonstrates the results on SVHN dataset. 
AT-GIF can improve the test accuracy of AT by 3.46\% on ResNet-18 under PGD-20 attack.
This manifests that the GIF can indeed improve adversarial robustness on various datasets.
% be applied in different scenarios. 

\subsection{Ablation Studies}
\label{sec:ablation}
In this section, we conduct a series of ablation studies to further understand the effects of different components in AT-GIF.

\paragraph{Generating adversarial training data with CW attack.}
Besides PGD-10 attack, we also utilize the CW attack~\citep{Carlini017_CW} with 10 steps (CW-10) to generate adversarial training data (Eq. (\ref{eq:pgd})) on ResNet-18.
%For a fair comparison, we use CW attack with 10 steps (CW-10) to generate adversarial training data. %Other settings are the same as our original settings.
Table \ref{tbl:cw} reports the results of generating adversarial training data with PGD-10 (AT-GIF (PGD)) and CW-10 (AT-GIF (CW)).  AT-GIF (CW) increases the robust test accuracy by 6.08\% on PGD-20 attack but has a lower performance on CW attack. The results show that generating adversarial training data with CW attack is also promising. 
%The results motivate us to investigate the effect of generating adversarial data with CW attack in future works.

%\paragraph{Gener of the interpolation weight $\lambda$ sampled from different distributions.}

\begin{table}[!t]
\scriptsize %\addtolength{\tabcolsep}{-5pt}
\centering
\normalsize
\caption{Test accuracy of ResNet-18 on the SVHN dataset.}
\label{tbl:svhn}
%\resizebox{1.99\columnwidth}{!}
\setlength{\tabcolsep}{3.44mm}{
\begin{tabular}{c|ccc}
\hline

Defense & Natural &PGD-20 &  CW  \\
\hline
$\quad$ AT $\quad$ & \textbf{95.67} & 74.52  & 74.86  \\
%\hline
$\quad$ AT-GIF $\quad$ & 94.98 & \textbf{77.98} & \textbf{75.10}  \\
\hline
\end{tabular}
}
%\vspace{-15pt}
\end{table}

\begin{table}[t!]
\scriptsize %\addtolength{\tabcolsep}{-5pt}
\centering
\normalsize
\caption{Test accuracy of AT-GIF with different methods of generating adversarial training data on the CIFAR-10 dataset.}
\label{tbl:cw}
%\resizebox{1.99\columnwidth}{!}
\setlength{\tabcolsep}{3.34mm}{
\begin{tabular}{c|ccc}
\hline
Defense & Natural &PGD-20  & CW\\
\hline
AT-GIF (PGD) & \textbf{81.59} & 53.57  & \textbf{50.07} \\
%\hline
AT-GIF (CW) & 81.19 & \textbf{59.65} & 47.79 \\
\hline
\end{tabular}
}
%\vspace{-15pt}
\end{table}

\begin{figure}[h!]
	\centering
	\subfigure{
		\includegraphics[width=0.38\linewidth]{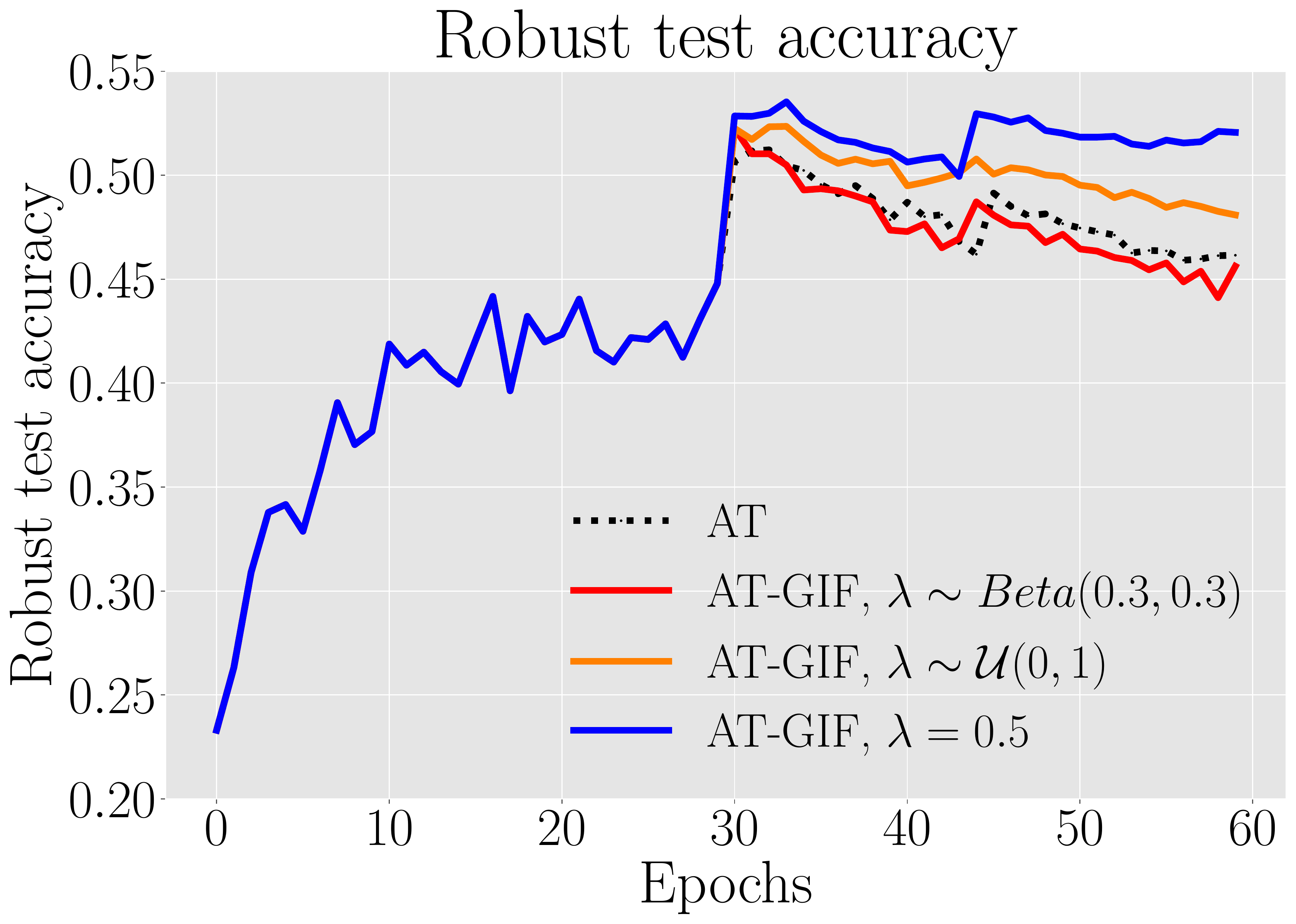}
	}
	\caption{Robust test accuracy of AT-GIF with different $\lambda$ under PGD-20 attack.}
	%\vspace{-3mm}
	\label{fig:weight}
\end{figure}

\paragraph{Selection of the interpolation weight $\lambda$.}
%To verify our fixed interpolation weight $\lambda=0.5$ can better benefit AT, 
We report the robustness of our fixed $\lambda=0.5$ with a random $\lambda$ sampled from a uniform distribution or a beta distribution, i.e., $\lambda\sim \mathcal{U}(0,1)$ or $\lambda\sim Beta(0.3,0.3)$ in Figure \ref{fig:weight}. The hyperparameters for uniform distribution and beta distribution are set according to~\citep{zhang2018mixup}.
%Figure \ref{fig:weight} shows the robust test accuracy of AT-GIF with $\lambda=0.5$, $\lambda\sim \mathcal{U}(0,1)$, and $\lambda\sim Beta(0.3,0.3)$ evaluated by PGD-20 attack. 
When $\lambda=0.5$ (blue line), the model achieves the best performance, which shows that $\lambda=0.5$ can indeed improve robustness. 
%When $\lambda\sim Beta(0.3,0.3)$ (red line), the model behaves similarly to standard AT (black dotted line). This is because $\lambda$ is close to 0 and 1, and the interpolated data have little difference with the original data. Thus, the interpolated data have little contribution to adversarial training.
The model with $\lambda\sim \mathcal{U}(0,1)$ (orange line) underperforms our AT-GIF with a fixed $\lambda=0.5$, due to that the uniform distribution of $\lambda$ leads to undesirable linear behavior. 
%The experiment further verifies our fixed interpolation weight $\lambda=0.5$ can mitigate the linear behavior between classes and benefit AT.
%This further verifies linear behavior leads to poor adversarial robustness and 
%\subsubsection{Different ratio of original data and interpolated data}

In addition, we report the robustness of AT-GIF with different fixed interpolation weight $\lambda\in\{0.1, 0.2, \ldots, 0.5\}$ on the CIFAR-10 dataset in Appendix \ref{app_sec:lambda}. 
%Note that $\lambda=0.1,\ldots,0.4$ is equivalent to $\lambda=0.9,\ldots,0.6$.
The results show that fixing $\lambda=0.5$ achieves the best robustness, which further verifies the GIF indeed aids in AT.

\paragraph{The effects of the burn-in period.} We conduct an experiment to show the effects of the burn-in period on ResNet-18.
Table \ref{tbl:burnin} compares the performance of AT-GIF with and without the burn-in period. AT-GIF (burn-in) trains with the original data without the need of the interpolated data in the first 30 epochs, and start to train with both the original data and the interpolated data after Epoch 30. AT-GIF (naive) utilizes the original data and the interpolated data for adversarial training from the first epoch. The results suggest that the burn-in period can slightly improve the robustness under CW attack, but has a slightly lower test accuracy under PGD-20 attack. 

\begin{table}[h!]
\scriptsize %\addtolength{\tabcolsep}{-5pt}
\centering
\normalsize
\caption{Test accuracy of AT-GIF with and without the burn-in period on the CIFAR-10 dataset.}
\label{tbl:burnin}
%\resizebox{1.99\columnwidth}{!}
\setlength{\tabcolsep}{2.9mm}{
\begin{tabular}{c|ccc}
\hline
Defense & Natural &PGD-20  & CW\\
\hline
AT-GIF (naive) & \textbf{82.35} & \textbf{53.60}  & 49.13 \\
%\hline
AT-GIF (burn-in) & 81.59 & 53.57 & \textbf{50.07} \\
\hline
\end{tabular}
}
%\vspace{-15pt}
\end{table}

\paragraph{Different ratios of original data and interpolated data.}
%In our experiments, the ratio of original data and interpolated data is 1:1. 
We train AT-GIF with different ratios of data from the original dataset and the interpolation set (e.g., 1:1, 1:2, and 2:1). To make sure these methods train with the same amount of data in one mini-batch, in AT-GIF (1:1), we set original batch size $m=64$ and  interpolation batch size $m'=64$; in AT-GIF (1:2), we set $m=43$ and $m'=85$; and in AT-GIF (2:1), we set $m=85$ and $m'=43$.
We report the test accuracy of these models in Appendix \ref{app_sec:ratio}. AT-GIF (1:1) has a better performance than AT-GIF (1:2) and AT-GIF (2:1). 
This indicates that selecting an appropriate ratio of the original data and the attackable data is also important for robustness enhancement.
% In Sec. [ref], we train 
% [more details]

% 1. robust test accuracy of resnet18 on cifar-10.

% 2. robust test accuracy of wrn32-10 on cifar-10.

% 3. robust test accuracy of resnet18 on svhn.

% 4. robust test accuracy of fgsm on cifar-10.

% 5. different $\lambda$ (0.5, uniform, beta).

% 6. cw attack

% 7.hard label issue

\section{Conclusion}
This paper has proposed a novel data augmentation framework, i.e., guided interpolation framework (GIF), for adversarial training. The GIF utilizes the meta information of the previous epoch to guide the data interpolation process.
GIF can obtain a high ratio of attackable data by interpolation without sacrificing much the local invariance, efficiently enhancing adversarial robustness. Extensive experiments show our GIF boosts the robustness of AT. 
A promising future direction is to investigate the different impacts of attackable data and guarded data in AT.
%A promising future direction is to investigate the different contribution of the original data and the interpolated data in adversarial training.

%A promising future direction is to investigate how to better generate adversarial variants for interpolated data.
%and provide a large amount of attackable data. 

% \begin{algorithm}[!t]
%     \caption{Generate interpolation set}
%     \label{alg:interpolate}
%     %\hspace*{\algorithmicindent} 
%     \textbf{Input:} attackable set $A$, data number $n$ \\
%     \textbf{Output:} interpolation set $\overline{S}$
%     \begin{algorithmic} % The number tells where the line numbering should start
%     \Procedure{IntSetGen}{}%{$\mathcal{S}$}
%         \State $\overline{S}=\emptyset$
%         \For{$i=1,..., n$}
%             \State Randomly sample two data $(x_{i,1},y_{i,1}), (x_{i,2},y_{i,2})$ from $A$
%             \State $\overline{x}_i\gets 0.5*x_{i,1}+0.5*x_{i,2}$
%             \State $\overline{y}_i\gets 0.5*y_{i,1}+0.5*y_{i,2}$
%             \State $\overline{S}\gets\overline{S}\cup{(\overline{x}_i,\overline{y}_i)}$
%         \EndFor
%     \EndProcedure
%     \end{algorithmic}
% \end{algorithm}

\section{Acknowledgement}
This work was supported by the Key Research and Development Program of Zhejiang Province of China (No. 2020C01024), and the National Natural Science Foundation of China under Grant No. 62050099, and the Natural Science Foundation of Zhejiang Province of China (No. LY18F020005).
JZ, GN, and MS were supported by JST AIP Acceleration Research Grant Number JPMJCR20U3, Japan. MS was also supported by the Institute for AI and Beyond, UTokyo.

\clearpage
\bibliographystyle{unsrtnat}  
\bibliography{references}  %%% Remove comment to use the external .bib file (using bibtex).

\appendix
\onecolumn

\section{Algorithm of TRADES-GIF, GAIRAT-GIF, and FastAT-GIF}
\label{app_sec:alg}
To show our GIF is a compatible method, we modify TRADES, GAIRAT, and FastAT to their GIF versions, i.e., TRADES-GIF, GAIRAT-GIF, and Fast-GIF.
The training of TRADES-GIF is shown in Algorithm \ref{app_alg:TRADES-GIF}. $\ell_{CE}$ is the cross-entropy loss, $\ell_{KL}$ is the Killback-Leibler loss, $\mathcal{N}(0,\mathbf{I})$ is the Gaussian distribution with zero mean and identity variance, $\gamma>0$ is a small constant controlling the strength of the Gaussian distribution, and $\beta>0$ is the weight of the KL loss.
The training of GAIRAT-GIF is shown in Algorithm \ref{app_alg:GAIRAT-GIF}. $\kappa(x,y)$ is the least iteration number that the PGD method need to generate adversarial variant $\tilde{x}$ to fool the network, and $\omega(x,y)$ is the geometry-aware weight assignment function.
The training of FastAT-GIF is shown in Algorithm \ref{app_alg:FastAT-GIF}. $\mathcal{U}(\cdot,\cdot)$ is the uniform distribution.

\begin{algorithm}[h!]
    \caption{TRADES with guided interpolation framework (TRADES-GIF)}
    \label{app_alg:TRADES-GIF}
    \textbf{Input:} model $f_\theta$, training dataset $S$, learning rate $\eta$, number of epochs $T$, original batch size $m$, interpolation batch size $m'$, number of batches $M$, perturbation bound $\epsilon$, step size $\alpha$, PGD step number $K$, KL loss weight $\beta$, and Gaussian noise weight $\gamma$ \\
    \textbf{Output:} adversarially robust network $f_\theta$
    \begin{algorithmic} 
        \State $A_0\gets S$
        \For{$\text{epoch } t= 1, 2, ..., T$}
            \State $A_t=\emptyset$
            \State Compute interpolation set $\overline{S}_t$ by $A_{t-1}$ and Eq. (\ref{eq:gif})
            \For{$\text{mini-batch }=1,..., M$}
                \State Randomly sample $\{(x_i,y_i)\}_{i=1}^{m}$ from $S$ and $\{(x_i,y_i)\}_{i=m+1}^{m+m'}$ from ${\overline{S}_t}$
                \For{$i=1, \ldots, m+m'$} {\textbf{in parallel}}  
                    %\State Generate adversarial data $\widetilde{x}_i$ of $x_i$ by Eq. (\ref{eq:pgd})
                    \State $\widetilde{x}_i\gets x_i+\gamma\mathcal{N}(0,\mathbf{I})$
                    \For{$k=1, \ldots, K$}
                        \State $\widetilde{x}_i\gets\Pi_{\mathcal{B}_\epsilon(x_i)}\left(\alpha \text{sign}\left(\nabla_{\widetilde{x}_i}\ell_{KL}(f_\theta(\widetilde{x}_i),f_\theta(x_i))\right)+\widetilde{x}_i\right)$
                    \EndFor
                    \If{$i\leq m$ \textbf{and} $\argmax_c f^c_\theta(\tilde{x}_i)\neq\argmax_c y_i^c$}
                    \State $A_t\gets A_t\cup\{(x_i,y_i)\}$
                    \EndIf
                \EndFor
                % \State Randomly sample $\{(x_i,y_i)\}_{i=m+1}^{m+m'}$ from ${\overline{S}_t}$
                % \For{$i=m+1,...,m+m'$} {\textbf{in parallel}}
                %     %\State Generate adversarial data $\widetilde{x}_i$ of $x_i$ by Eq. (\ref{eq:pgd})
                %     \State $\widetilde{x}_i\gets x_i+\gamma\mathcal{N}(0,I)$
                %     \For{$k=1, \ldots, K$}
                %         \State $\widetilde{x}_i\gets\Pi_{\mathcal{B}_\epsilon(x_i)}\left(\alpha \text{sign}\left(\nabla_{\widetilde{x}_i}\ell_{KL}(f_\theta(\widetilde{x}_i),f_\theta(x_i))\right)+\widetilde{x}_i\right)$
                %     \EndFor
                % \EndFor
                \State $\theta\gets\theta-\eta\cdot\frac{1}{m+m'}\sum_{i=1}^{m+m'} \nabla_\theta[\ell_{CE}(f_\theta(\widetilde{x}_i),y_i)+\beta\ell_{KL}(f_\theta(\widetilde{x}_i),y_i)]$
            \EndFor
        \EndFor
    \end{algorithmic}
\end{algorithm}

\begin{algorithm}[h!]
    \caption{GAIRAT with guided interpolation framework (GAIRAT-GIF)}
    \label{app_alg:GAIRAT-GIF}
    \textbf{Input:} model $f_\theta$, training dataset $S$, learning rate $\eta$, number of epochs $T$, original batch size $m$, interpolation batch size $m'$, number of batches $M$, perturbation bound $\epsilon$, step size $\alpha$, PGD step number $K$ \\
    \textbf{Output:} adversarially robust network $f_\theta$
    \begin{algorithmic} 
        \State $A_0\gets S$
        \For{$\text{epoch } t= 1, 2, ..., T$}
            \State $A_t=\emptyset$
            \State Compute interpolation set $\overline{S}_t$ by $A_{t-1}$ and Eq. (\ref{eq:gif})
            \For{$\text{mini-batch }=1,..., M$}
                \State Randomly sample $\{(x_i,y_i)\}_{i=1}^{m}$ from $S$ and $\{(x_i,y_i)\}_{i=m+1}^{m+m'}$ from ${\overline{S}_t}$
                \For{$i=1, \ldots, m+m'$} {\textbf{in parallel}}
                    %\State Generate adversarial data $\widetilde{x}_i$ of $x_i$ by Eq. (\ref{eq:pgd}) and calculate $\omega(x_i,y_i)$ according to GAIRAT
                    \State $\widetilde{x}_i\gets x_i$; $\kappa(x_i,y_i)\gets 0$
                    \For{$k=1, \ldots, K$}
                        \If{ $\tilde{x}$ is predicted correctly}
                            \State $\kappa(x_i,y_i)\gets\kappa(x_i,y_i)+1$
                        \EndIf
                        \State $\widetilde{x}_i\gets\Pi_{\mathcal{B}_\epsilon(x_i)}\left(\alpha \text{sign}\left(\nabla_{\widetilde{x}_i}\ell(f_\theta(\widetilde{x}_i),y_i)\right)+\widetilde{x}_i\right)$
                    \EndFor
                    \State Compute $\omega(x_i,y_i)$ according to $\kappa(x_i,y_i)$
                    \If{$i\leq m$ \textbf{and} $\argmax_c f^c_\theta(\tilde{x}_i)\neq\argmax_c y_i^c$}
                    \State $A_t\gets A_t\cup\{(x_i,y_i)\}$
                    \EndIf
                \EndFor
                % \State Randomly sample $\{(x_i,y_i)\}_{i=m+1}^{m+m'}$ from ${\overline{S}_t}$
                % \For{$i=m+1,...,m+m'$} {\textbf{in parallel}}
                %     \State Generate adversarial data $\widetilde{x}_i$ of $x_i$ by Eq. (\ref{eq:pgd})
                %     and calculate $\omega(x_i,y_i)$ according to GAIRAT
                % \EndFor
                %\State $\theta\gets\theta-\eta\cdot \nabla_\theta \left\{\sum_{i=1}^{m}\frac{\omega(x_i,y_i)\ell(f_\theta(\widetilde{x}_i),y_i)}{\sum_{i=1}^{m}\omega(x_i,y_i)+\sum_{i=m+1}^{m+m'}\omega(\overline{x}_i,\overline{y}_i)} + \sum_{i=m+1}^{m+m'}\frac{\omega(\overline{x}_i,\overline{y}_i)\ell(f_\theta(\widetilde{x}_i),\overline{y}_i)}{\sum_{i=1}^{m}\omega(x_i,y_i)+\sum_{i=m+1}^{m+m'}\omega(\overline{x}_i,\overline{y}_i)}\right\}$
                \State $\theta\gets\theta-\eta\cdot \nabla_\theta \left\{\sum_{i=1}^{m+m'}\frac{\omega(x_i,y_i)}{\sum_{j=1}^{m+m'}\omega(x_j,y_j)}\ell(f_\theta(\widetilde{x}_i),y_i)\right\}$
            \EndFor
        \EndFor
    \end{algorithmic}
\end{algorithm}

\begin{algorithm}[h!]
    \caption{FastAT with guided interpolation framework (FastAT-GIF)}
    \label{app_alg:FastAT-GIF}
    \textbf{Input:} model $f_\theta$, training dataset $S$, learning rate $\eta$, number of epochs $T$, original batch size $m$, interpolation batch size $m'$, number of batches $M$, perturbation bound $\epsilon$, step size $\alpha$, PGD step number $K$ \\
    \textbf{Output:} adversarially robust network $f_\theta$
    \begin{algorithmic} 
        \State $A_0\gets S$
        \For{$\text{epoch } t= 1, 2, ..., T$}
            \State $A_t=\emptyset$
            \State Compute interpolation set $\overline{S}_t$ by $A_{t-1}$ and Eq. (\ref{eq:gif})
            \For{$\text{mini-batch }=1,..., M$}
                \State Randomly sample $\{(x_i,y_i)\}_{i=1}^{m}$ from $S$ and $\{(x_i,y_i)\}_{i=m+1}^{m+m'}$ from ${\overline{S}_t}$
                \For{$i=1, \ldots, m+m'$} {\textbf{in parallel}}
                    %\State Generate adversarial data $\widetilde{x}_i$ of $x_i$ by FastAT
                    \State $\widetilde{x}_i\gets x_i+\mathcal{U}(-\epsilon,\epsilon)$
                    \State $\widetilde{x}_i\gets\Pi_{\mathcal{B}_\epsilon(x_i)}\left(\alpha \text{sign}\left(\nabla_{\widetilde{x}_i}\ell(f_\theta(\widetilde{x}_i),y_i)\right)+\widetilde{x}_i\right)$
                    \If{$i\leq m$ \textbf{and} $\argmax_c f^c_\theta(\tilde{x}_i)\neq\argmax_c y_i^c$}
                    \State $A_t\gets A_t\cup\{(x_i,y_i)\}$
                    \EndIf
                \EndFor
                % \State Randomly sample $\{(x_i,y_i)\}_{i=m+1}^{m+m'}$ from ${\overline{S}_t}$
                % \For{$i=m+1,...,m+m'$} {\textbf{in parallel}}
                %     %\State Generate adversarial data $\widetilde{x}_i$ of $x_i$ by FastAT
                %     \State $\widetilde{x}_i\gets x_i+\mathcal{U}(-\epsilon,\epsilon)$
                %     \State $\widetilde{x}_i\gets\Pi_{\mathcal{B}_\epsilon(x_i)}\left(\alpha \text{sign}\left(\nabla_{\widetilde{x}_i}\ell(f_\theta(\widetilde{x}_i),y_i)\right)+\widetilde{x}_i\right)$
                % \EndFor
                \State $\theta\gets\theta-\eta\cdot\frac{1}{m+m'}\sum_{i=1}^{m+m'} \nabla_\theta\ell(f_\theta(\widetilde{x}_i),y_i)$
            \EndFor
        \EndFor
    \end{algorithmic}
\end{algorithm}
\clearpage
\section{Detailed Experimental Setttings}
\subsection{Detailed Experimental Settings of Figure 1 and 2}
\label{app_sec:fig12}
In Figure 1, we train three models with AT, AT-mixup, and AT-GIF respectively. The settings for these models are the same as our default settings (described in Section 4), except these models do not involve the burn-in period. The training of AT-mixup is shown in Algorithm \ref{app_alg:mixup}, and the training of AT-GIF is shown in Algorithm 1.

In Figure 2, we randomly select 1,600 data from 2 random classes (800 data for each class) from the CIFAR-10 dataset. We embed the data to 2-dimension with PCA~\cite{abdi2010principal}. Then, we train these 2D data on a small neural network with four fully connected layers (size 10--10--5--2) with AT, AT-mixup, and AT-GIF, respectively. We visualize the prediction confidences in the three lower panels.

\begin{algorithm}[h!]
    %\setstretch{2.35}
    \caption{Adversarial training with vanilla mixup (AT-mixup)}
    \label{app_alg:mixup}
    %\hspace*{\algorithmicindent} 
    \textbf{Input:} model $f_\theta$, training dataset $S$, learning rate $\eta$, number of epochs $T$, original batch size $m$, interpolation batch size $m'$, number of batches $M$, perturbation bound $\epsilon$, step size $\alpha$, PGD step number $K$ \\
    \textbf{Output:} adversarially robust network $f_\theta$
    \begin{algorithmic} 
        \State $A_0\gets S$
        \For{$\text{epoch } t= 1, 2, ..., T$}
            \State $A_t=\emptyset$
            \State Compute interpolation set $\overline{S}_t$ by training dataset $S$ and Eq. (\ref{eq:mixup}) 
            \For{$\text{mini-batch }=1,..., M$}
                \State Randomly sample $\{(x_i,y_i)\}_{i=1}^{m}$ from $S$
                \For{$i=1, \ldots, m$} {\textbf{in parallel}} 
                    \State Generate adversarial data $\widetilde{x}_i$ of $x_i$ by Eq. (\ref{eq:pgd}) 
                \EndFor
                \State Randomly sample $\{(x_i,y_i)\}_{i=m+1}^{m+m'}$ from ${\overline{S}_t}$
                \For{$i=m+1,...,m+m'$} {\textbf{in parallel}}
                    \State Generate adversarial data $\widetilde{x}_i$ of $x_i$ by Eq. (\ref{eq:pgd}) 
                \EndFor
                \State $\theta\gets\theta-\eta\cdot\frac{1}{m+m'}\sum_{i=1}^{m+m'} \nabla_\theta\ell(f_\theta(\widetilde{x}_i),y_i)$
                
            \EndFor
        \EndFor
    \end{algorithmic}
\end{algorithm}

\subsection{Detailed Description of the Datasets}
\label{app_sec:dataset}
In our experiments, all the models are trained on the CIFAR-10~\cite{krizhevsky2009learning_cifar10} and the SVHN~\cite{netzer2011reading_SVHN} dataset.

The CIFAR-10 dataset consists of 60,000 32x32 colour images in 10 classes, with 6,000 images per class. There are 50,000 training images and 10,000 test images.

SVHN is a real-world image dataset consists of 73,257 training data and 26,032 testing data in 10 classes. SVHN is obtained from house numbers in Google Street View images.
\subsection{Detailed Settings of Baselines}
\label{app_sec:baseline}
For fair comparisons, our GIF-based methods (TRADES-GIF, GAIRAT-GIF, and FastAT-GIF) keep the same settings as their original versions (TRADES~\cite{zhang2019theoretically}, GAIRAT~\cite{zhang2021geometry}, FastAT~\cite{wong2020fast_zico_kolter}). In TRADES and TRADES-GIF, we set KL loss weight $\beta=6$, weight of Gaussian noise $\gamma=0.001$, and training epoch $T=75$. The initial learning rate $\eta=0.1$ and divided by 10 at Epoch 37 and 56, respectively. In GAIRAT and GAIRAT-GIF, we set geometry-aware weight assignment function $\omega(x,y)=\frac{1+\text{tanh}(-1+5\times(1-2\times\kappa(x,y)/K))}{2}$ according to \cite{zhang2021geometry}.

\section{Additional Experiments}
\label{app_sec:add_exp}

\subsection{Auxiliary Experiment for Comparisons Between AT, AT-mixup, and AT-GIF}
\label{app_sec:aux_exp}
\begin{figure*}[h!]
  \centering
  \subfigure{
		\includegraphics[width=0.6\linewidth]{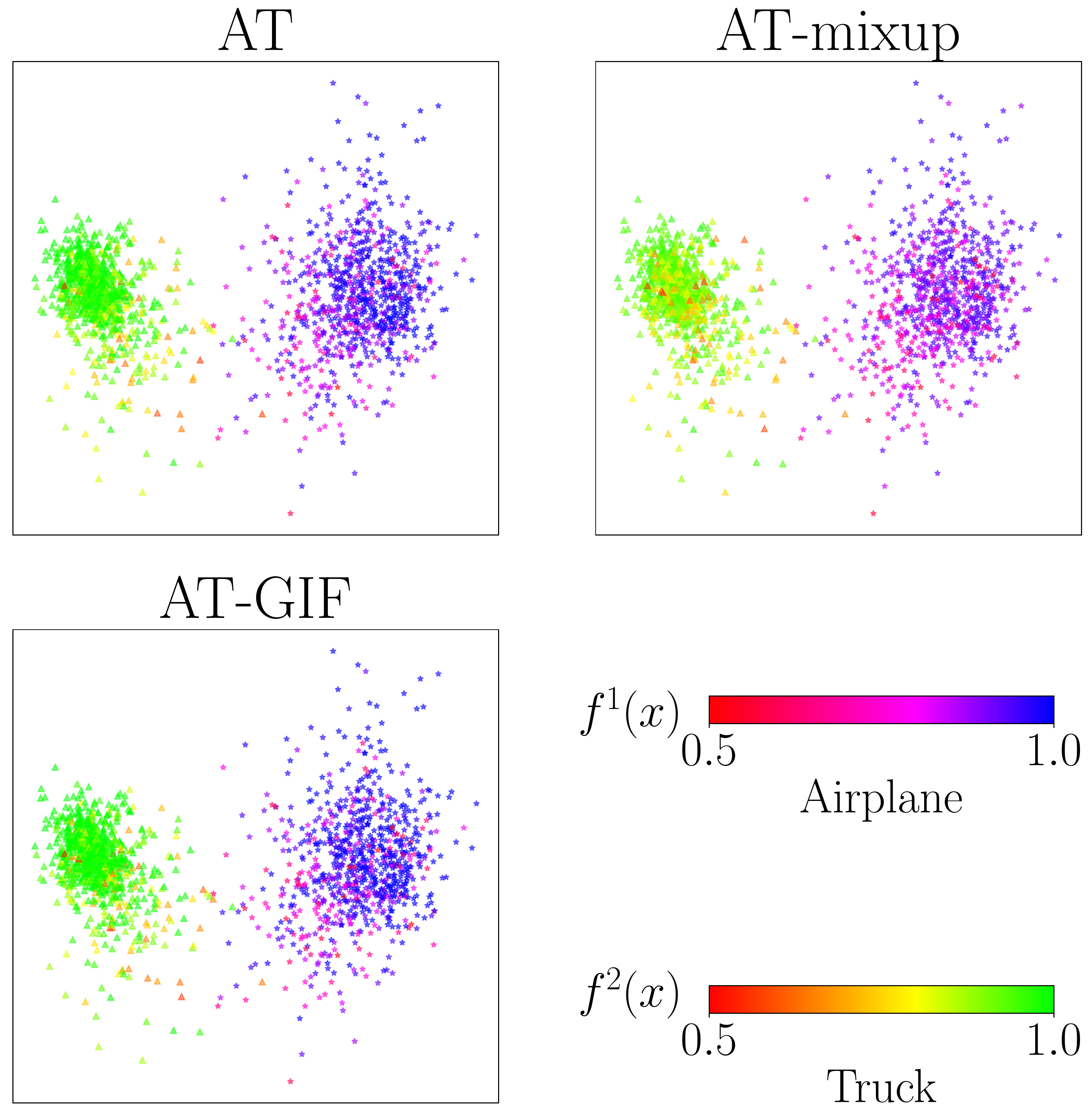}
	}
  %\vspace{-3mm}
  \caption{Comparisons between AT, AT-mixup, and AT-GIF on ResNet-18~\cite{he2016deep}.}
  \label{app_fig:mixup_bound}
  %\vspace{-4mm}
\end{figure*}

We conduct an auxiliary experiment to further compare the differences between AT, AT-mixup, and AT-GIF on ResNet-18~\cite{he2016deep}.
First, we pretrain three ResNet-18 models on the whole CIFAR-10 dataset with AT, AT-mixup, and AT-GIF, respectively.
Second, we randomly select 2,000 testing data from 2 random classes (1,000 data for each class) from the CIFAR-10 dataset. 
Third, we compute the prediction confidences of these 2,000 data with the three pretrained model. 
Fourth, we embed these data to 2-dimension with PCA~\cite{abdi2010principal}.
Last, we visualize these data according to their prediction confidence in Figure \ref{app_fig:mixup_bound}.
The prediction confidences of the model trained with AT-mixup behaves linearly.
AT-GIF alleviates the linear issue and behaves sharply.

\subsection{Robustness Evaluation of GIF Incorporating with Other AT methods on ResNet-18}
\label{app_sec:gif+others}
We report the test accuracy of GIF incorporating with other AT methods on ResNet-18 in Table \ref{app_tbl:gif+others}. The results show that all GIF-based methods outperform their original versions.

\begin{table*}[h!]
\scriptsize %\addtolength{\tabcolsep}{-5pt}
\centering
\normalsize
\caption{Test accuracy of GIF incorporating with other AT methods on ResNet-18.}
\label{app_tbl:gif+others}
%\resizebox{1.99\columnwidth}{!}
{
\begin{tabular}{c|c|cccc|cccc}
\hline
\multirow{2}*{Defense} & \multirow{2}*{Network} & \multicolumn{4}{c|}{Best} &\multicolumn{4}{c}{Last} \\
\cline{3-10}
& & Natural & PGD-20 &  CW &  AA & Natural & PGD-20 & CW & AA \\
\hline
TRADES & \multirow{2}*{ResNet-18} & 81.28 & 52.05  &  49.63 & 48.53  & 81.17  & 51.81 & 49.35 & 48.41\\
%\cline{3-10}
TRADES-GIF & &\textbf{81.54} & \textbf{53.04} & \textbf{49.84} & \textbf{48.96} & \textbf{82.11} & \textbf{52.05} & \textbf{49.51} & \textbf{48.45}  \\
\hline
GAIRAT & \multirow{2}*{ResNet-18} & \textbf{82.91} & 55.93 & 42.74  & 40.03 & \textbf{83.03}  & 50.94 & 38.04 & 35.17\\
GAIRAT-GIF &  & 81.29 & \textbf{60.57} & \textbf{43.07} &  \textbf{40.36} & 81.07 & \textbf{60.08} & \textbf{40.09} & \textbf{37.43}\\
\hline
Fast AT & \multirow{2}*{ResNet-18} & \textbf{86.33} & 43.81  & 45.12  &  41.03 & 89.26  & 1.44 & 2.72 & 0.00 \\
Fast AT-GIF & & 85.45 & \textbf{47.49} & \textbf{45.88}  & \textbf{42.88}  &  89.13 &  1.23 & 2.35 & 0.00 \\
\hline %\cellcolor{gray!30}
\end{tabular}
}
%\vspace{-15pt}
\end{table*}

\subsection{Robustness Evaluation of AT-GIF with Different Fixed Interpolation Weight $\lambda$}
\label{app_sec:lambda}
We report the test accuracy of AT-GIF with different fixed interpolation weight $\lambda\in\{0.1,0.2,0.3,0.4,0.5\}$. The results suggest that AT-GIF with $\lambda=0.5$ achieves the best performance.
\begin{table}[h!]
\scriptsize %\addtolength{\tabcolsep}{-5pt}
\centering
\normalsize
\caption{Test accuracy of AT-GIF with different fixed interpolation weight $\lambda$.}
\label{tbl:lambda}
%\resizebox{1.99\columnwidth}{!}
\begin{tabular}{c|ccccc}
\hline
Metric & AT-GIF (0.1) & AT-GIF (0.2)  & AT-GIF (0.3) &AT-GIF (0.4) & AT-GIF (0.5) \\
\hline
Natural & \textbf{84.42} & 84.10 & 83.73 & 83.46 & 81.59 \\
PGD-20 & 47.88 & 49.72 & 52.27  & 52.81 & \textbf{53.57} \\
CW & 46.85 & 48.30 & 48.88 & 49.08 &  \textbf{50.07} \\
\hline
\end{tabular}
%\vspace{-15pt}
\end{table}

\subsection{Robustness Evaluation of AT-GIF with Different Ratios of Original Data and Interpolated Data}
\label{app_sec:ratio}
We report the test accuracy of AT-GIF with different ratios of data from the original dataset and interpolation set on the CIFAR-10 dataset~\cite{krizhevsky2009learning_cifar10} in Table \ref{app_tbl:ratio}. The results suggest that GIF(1:1) is most effective for enhancing robustness.

\begin{table}[h!]
\scriptsize %\addtolength{\tabcolsep}{-5pt}
\centering
\normalsize
\caption{Test accuracy of AT-GIF with different ratios of original data and interpolated data on ResNet-18.}
\label{app_tbl:ratio}
%\resizebox{1.99\columnwidth}{!}
\setlength{\tabcolsep}{4.5mm}{
\begin{tabular}{c|ccc}
\hline
Defense & Natural &PGD-20  & CW\\
\hline
AT-GIF (1:1) & 81.59 & \textbf{53.57} & \textbf{50.07} \\
AT-GIF (1:2) & 81.90 & 53.05 & 49.33 \\
AT-GIF (2:1) & \textbf{82.88} & 51.98 & 49.47 \\
\hline
\end{tabular}
}
%\vspace{-15pt}
\end{table}
\end{document}